\documentclass[nohyperref]{article}

\usepackage{bm}
\usepackage{bbding}
\usepackage[table]{xcolor}
\usepackage{multirow}
\usepackage{enumerate}
\usepackage{microtype}
\usepackage{graphicx}
\usepackage{subfigure}
\usepackage{booktabs} 

\usepackage{hyperref}

\usepackage[accepted]{icml2022}

\usepackage{amsmath}
\usepackage{amssymb}
\usepackage{mathtools}
\usepackage{amsthm}

\usepackage[capitalize,noabbrev]{cleveref}

\theoremstyle{plain}

\theoremstyle{definition}

\theoremstyle{remark}

\usepackage[textsize=tiny]{todonotes}

\icmltitlerunning{Multiple Thinking Achieving Meta-Ability Decoupling for Object Navigation}

\begin{document}

\twocolumn[
\icmltitle{Multiple Thinking Achieving Meta-Ability Decoupling for Object Navigation}



\icmlsetsymbol{equal}{*}

\begin{icmlauthorlist}
\icmlauthor{Ronghao Dang}{tongji}
\icmlauthor{Lu Chen}{tongji}
\icmlauthor{Liuyi Wang}{tongji}
\icmlauthor{Zongtao He}{tongji}
\icmlauthor{Chengju Liu}{tongji}
\icmlauthor{Qijun Chen}{tongji}
\end{icmlauthorlist}

\icmlaffiliation{tongji}{Department of Control Science and Engineering, Tongji University, Shanghai 201804, China}

\icmlcorrespondingauthor{Ronghao Dang}{dangronghao@tongji.edu.cn}
\icmlcorrespondingauthor{Lu Chen}{chenlu\_i@tongji.edu.cn}
\icmlcorrespondingauthor{Liuyi Wang}{wly@tongji.edu.cn}
\icmlcorrespondingauthor{Zongtao He}{xingchen327@tongji.edu.cn}
\icmlcorrespondingauthor{Chengju Liu}{liuchengju@tongji.edu.cn}
\icmlcorrespondingauthor{Qijun Chen}{qjchen@tongji.edu.cn}
\icmlkeywords{Machine Learning, ICML}

\vskip 0.3in
]



\begin{abstract}
We propose a meta-ability decoupling (MAD) paradigm, which brings together various object navigation methods in an architecture system, allowing them to mutually enhance each other and evolve together.
Based on the MAD paradigm, we design a multiple thinking (MT) model that leverages distinct thinking to abstract various meta-abilities. Our method decouples meta-abilities from three aspects: input, encoding, and reward while employing the multiple thinking collaboration (MTC) module to promote mutual cooperation between thinking. MAD introduces a novel qualitative and quantitative interpretability system for object navigation. Through extensive experiments on AI2-Thor and RoboTHOR, we demonstrate that our method outperforms state-of-the-art (SOTA) methods on both typical and zero-shot object navigation tasks.  
\end{abstract}

\section{Introduction}

\begin{figure}[t]
\begin{center}
\centerline{\includegraphics[width=\columnwidth]{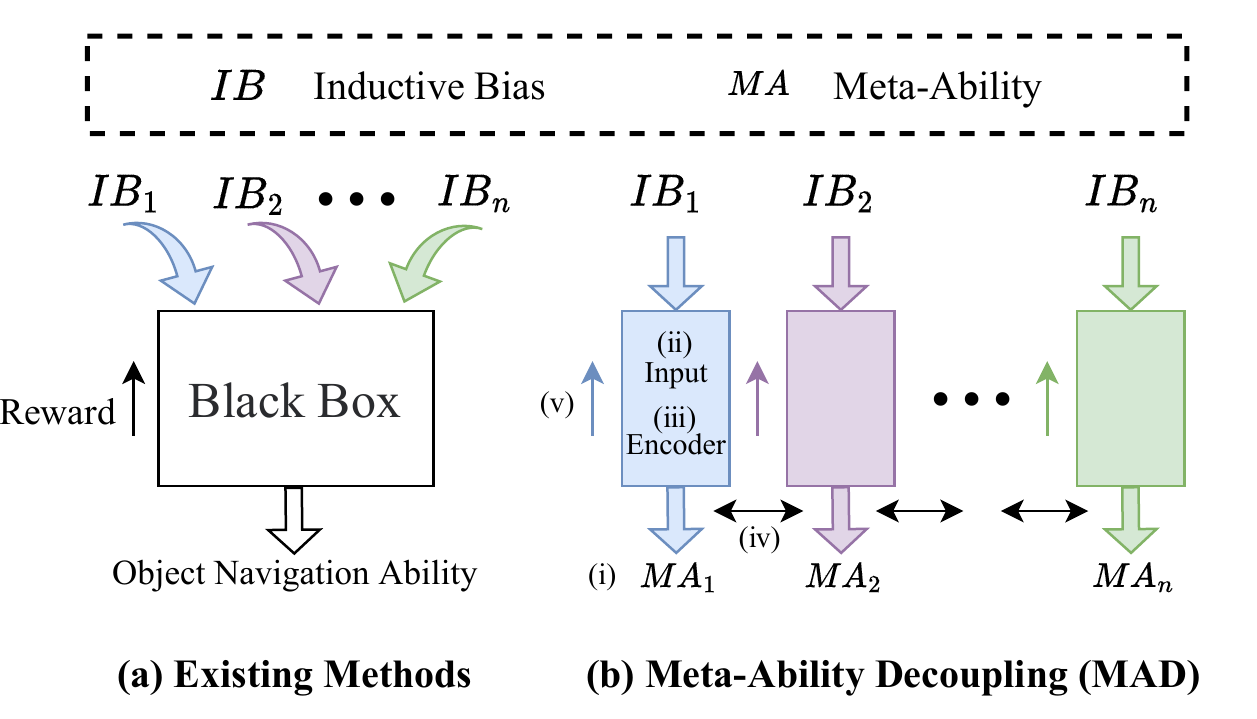}}
\caption{(a) Existing methods directly improve the overall object navigation ability by introducing various inductive biases into the black box model. (b) Our proposed meta-ability decoupling (MAD) paradigm decomposes the overall object navigation ability into multiple meta-abilities, and designs specific inputs, thinking encoders, and rewards for each meta-ability.}
\label{compare_method_model}
\end{center}
\vskip -0.2in
\end{figure}

Object navigation \cite{COS-POMDP, ForeSI, VTNet, RES-StS} is a challenging task that requires an agent to find a target object in an unknown environment with first-person visual observations. 
Numerous techniques have been developed to advance this field by incorporating different inductive biases (Figure~\ref{compare_method_model} (a)) due to the task's complexity.
However, regrettably, the object navigation field does not form a unified inductive bias paradigm similar to the CV \cite{Tobias, convit} or NLP \cite{NLM_IB, sq2sq_IB} fields. Inspired by the flaw, through the induction and sublimation of the current mainstream methods, we propose a meta-ability decoupling (MAD) paradigm, hoping to unify and connect various object navigation methods. 

This paper involves two important new concepts: meta-ability and thinking. \textbf{Meta-ability} refers to every essential ability needed to complete a complex task. 
For instance, solving a mathematical problem requires the integration of various meta-abilities such as text comprehension, logical reasoning, and conceptual abstraction. Without these meta-abilities, relying solely on intuition is insufficient to complete complex tasks.
\textbf{Thinking} refers to the information abstraction for a certain ability. Typically, this abstraction is modeled end-to-end using neural networks. 

\begin{figure}[t]
\begin{center}
\centerline{\includegraphics[width=0.8\columnwidth]{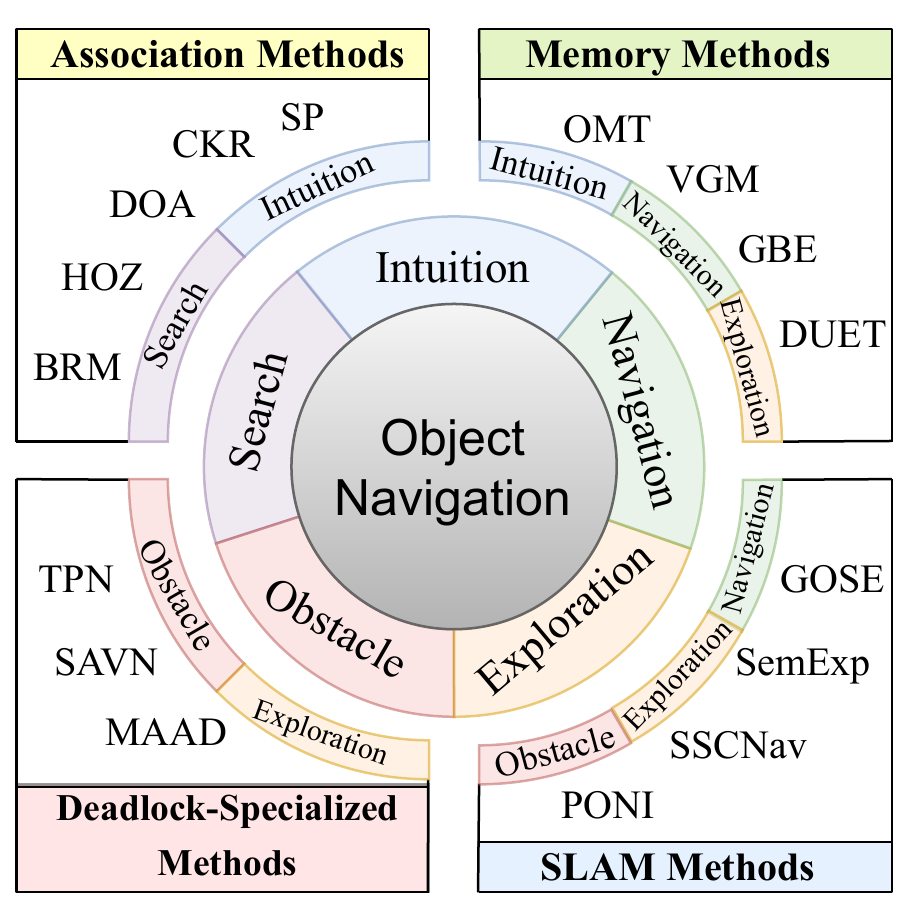}}
\caption{\textbf{Summary of various object navigation methods}. We categorize the mainstream methods for object navigation into four classes, which achieve the enhancement of certain meta-abilities by improving the neural network.}
\label{related_works}
\end{center}
\vskip -0.2in
\end{figure}

According to the definition of meta-ability and thinking, we summarize the current mainstream object navigation methods and identify their limitations. As shown in Figure~\ref{related_works}, object navigation methods are divided into four categories: association methods \cite{DOA, HOZ}, memory methods \cite{DUET, OMT}, deadlock-specialized methods \cite{TPN, MAAD} and SLAM methods \cite{SemExp, SSCNav}. The different inductive biases introduced by these four types of methods determine which meta-abilities are emphasized and which are overlooked. Therefore, the existing methods all attempt to use biased thinking to abstract the ultimate ability for object navigation (Figure~\ref{compare_method_model} (a)). Nevertheless, due to the sparsity and ambiguity of the reward signal, it is challenging for biased thinking to implicitly decouple complete meta-abilities which are crucial in object navigation.

To address the above issues, we propose a meta-ability decoupling (MAD) paradigm (Figure~\ref{compare_method_model} (b)), which solves embodied AI tasks in five stages: (\romannumeral1) selecting meta-abilities based on prior knowledge; (\romannumeral2) determining the input features of each thinking according to the characteristics of its corresponding meta-ability; (\romannumeral3) designing suitable encoding networks for each thinking; (\romannumeral4) designing the collaboration modules between different thinking according to the characteristics of the task;
(\romannumeral5) designing rewards and punishments for each meta-ability. During this process, meta-abilities are decoupled in three aspects: input, encoding, and reward signals.  
In this paper, we primarily focus on the investigation of object navigation tasks, however, we believe that the MAD paradigm can be extended to other similar embodied AI tasks.

Guided by the MAD paradigm, we design a multiple thinking (MT) model for the object navigation task. First, we select five meta-abilities (explained in Sec.~\ref{sec:MAD}): intuition, search, navigation, exploration and obstacle. Subsequently, for these five meta-abilities, we use overall image features, object detection features, target-oriented memory, historical state memory, and obstacle location memory as input for corresponding thinking. Each thinking uses a simple encoding network with necessary inductive bias. 
Furthermore, we devise a multiple thinking collaboration (MTC) module to facilitate cooperation between the different meta-abilities.
Finally, meta-ability reward is designed to guide each thinking's abstract understanding for the corresponding meta-ability.

Extensive experiments on the AI2-Thor \cite{AI2-THOR} and RoboTHOR \cite{RoboTHOR} datasets show that our MAD paradigm not only outperforms SOTA methods on the typical object navigation task, but also on the zero-shot object navigation task. Moreover, an interpretability analysis of MT model based on MAD demonstrates that our method contributes significantly to both the interpretability and flexibility of object navigation tasks. Our contributions can be summarized as follows:

\begin{itemize}
\item We propose a general meta-ability decoupling (MAD) paradigm to generalize and unify various current object navigation approaches.
\item Following the MAD paradigm, we design a multiple thinking (MT) model for the object navigation task, which outperforms existing models in both typical and zero-shot object navigation tasks.
\item Our meta-ability interpretability framework provides a novel analytical mode for future researchers.
\end{itemize}

\section{Related Works}

\subsection{Object Navigation}
\textbf{Target-specific typical object navigation tasks} require an agent to navigate to a known target object in an unknown environment. Some recent methods diligently improve the network or introduce prior knowledge in order to solve various problems in the object navigation task. We categorize these methods into four classes (Figure~\ref{related_works}): (\romannumeral1) association methods \cite{BRM, CKR, SP} which utilize object association or area association to enable the agent to build a relational graph model of the scene; (\romannumeral2) memory methods \cite{GBE, VGM} which depend on long-term explicit memory to more comprehensively consider historical information to make decisions; (\romannumeral3) SLAM methods \cite{GOSE, PONI} which build an agent-centric semantic map in real time; (\romannumeral4) deadlock-specialized methods \cite{SAVN, TPN} which use special mechanisms to help the agent escape from the local deadlock state.  Due to the lack of the meta-ability decoupling perspective, each class of methods only emphasize partial meta-abilities, resulting in a lack of comprehensive ability to solve complex tasks.

\textbf{Target-agnostic zero-shot object navigation tasks} are gaining increasing attention with the development of multimodal contrastive learning \cite{CLIP}. This task requires that the training environment shields the target objects for testing. Al-Halah et al. \citeyearpar{ZSEL}
mapped various modalities into the image-goal embedding space, thus adapting the image-goal navigation agent. 
Zhao et al. \citeyearpar{COS_ZSON} represented the object-target relationship as cosine similarity to alleviate the overfitting.
These zero-shot object navigation methods essentially extend the typical object navigation architecture by mapping the discrete class inputs to a continuous semantic space. 

\subsection{Decoupling Idea in Object Navigation}
Decoupling is a common concept in the field of artificial intelligence \cite{decouple_interpretable}, and it is also frequently observed in object navigation tasks. SemExp \cite{SemExp} decouples the continuous decision-making process into two discretized steps that are "where to look for an object?" and "how to navigate to (x, y)?". AVSW \cite{AVSW} decouples the environmental exploration from navigation to the target. ANS \cite{ANS} decouples the modeling of the environment from the end-to-end network using a real-time built semantic map. The aforementioned decoupling methods all split the end-to-end object navigation model into several independently trained models, thus losing the advantages of flexibility and simplicity. Our MAD paradigm improves the interpretability and generalizability of the model while maintaining end-to-end learning. 

\section{Meta-Ability Decoupling (MAD)}
\label{sec:MAD}
Object navigation is a complex long-distance decision-making task in the real world. Efficient and accurate navigation to the target object requires the assistance of multiple meta-abilities. We decouple the following five meta-abilities from the object navigation task based on existing object navigation methods and human experience: 
\begin{enumerate}[1)]
\setlength{\itemsep}{-0.1cm}
\item \textbf{Intuition Ability:} The ability to directly derive action decisions from raw image features.

\item \textbf{Search Ability:} The ability to look for the target object through knowledge association. 

\item \textbf{Navigation Ability:} The ability to navigate to the target position based on the target orientation information in memory.

\item \textbf{Exploration Ability:} The ability to efficiently and comprehensively acquire scene information.

\item \textbf{Obstacle Ability:} The ability to avoid colliding with obstacles.

\end{enumerate}
All five meta-abilities are present in current methods (Figure~\ref{related_works}), however, a single method only emphasizes certain meta-abilities. By clearly identifying and decoupling them, researchers can more easily combine the strengths of multiple approaches.

\section{Multiple Thinking (MT) Network}

\begin{figure*}[t]
\begin{center}
\centerline{\includegraphics[width=\textwidth]{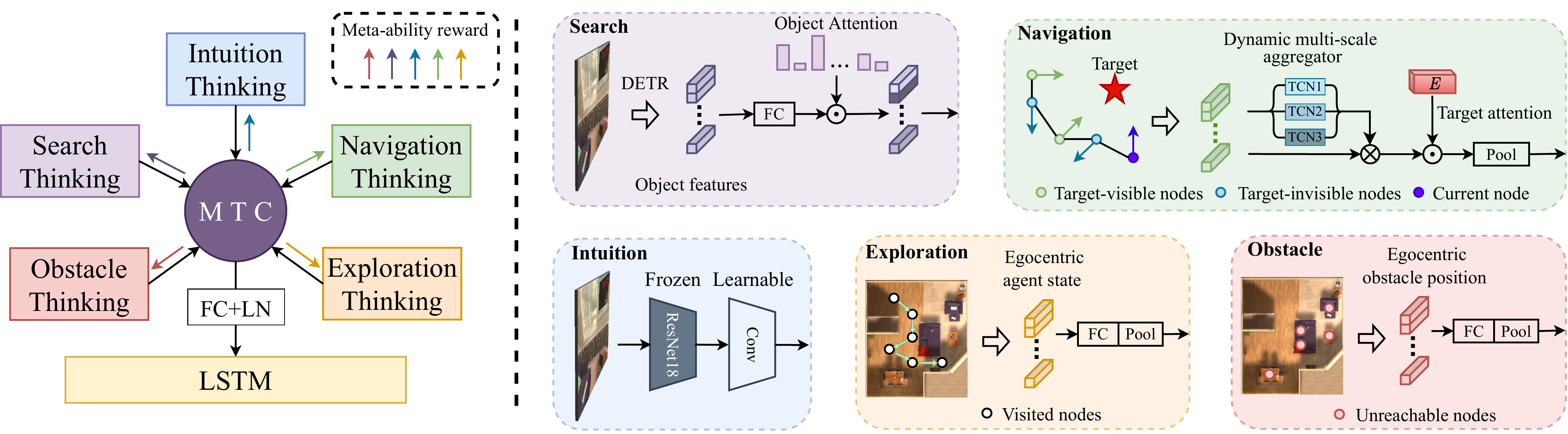}}
\caption{\textbf{Model overview}. MTC: multiple thinking collaboration. Our multiple thinking (MT) model is primarily composed of five thinking modules and a MTC module, preceding the LSTM network. The colored boxes on the right match the details of thinking embedding on the left. (1) Intuition thinking takes image ResNet18 encoding as input. (2) Search thinking takes the object features extracted by DETR as input. (3) Navigation thinking takes the target orientation information at target-visible nodes as input. (4) Exploration thinking takes historical agent states as input. (5) Obstacle thinking takes the locations of known unreachable nodes as input.}
\label{architecture}
\end{center}
\vskip -0.2in
\end{figure*}

The multiple thinking (MT) network is designed under the guidance of the MAD paradigm. As shown in Figure~\ref{architecture}, we design suitable input (Sec.~\ref{sec:Thinking Inputs}), encoding network (Sec.~\ref{sec:Thinking Embedding}), and reward (Sec.~\ref{sec:Meta-Ability Reward}) for each meta-ability. Additionally, we design a multiple thinking collaboration (MTC) module (Sec.~\ref{sec:MTC}) that interacts information between different types of thinking. 

\subsection{Task Definition}

The agent is initialized to a random state $ s=\{x,y,\theta,\beta \}$ and random target object $p$. At each timestamp $t$, according to the single view RGB image $o_t$ and target $p$, the agent learns a navigation strategy $\pi(a_t|o_t, p)$, where $a_t\in A=\{MoveAhead$; $RotateLeft$; $RotateRight$;  $LookDown$;  $LookUp$; $Done\}$ and $Done$ is the output if the agent believes that it has navigated to the target location. Ultimately, if the agent is within a threshold (i.e., 1.5 meters \cite{TPN}) of the target and correctly detects it when $Done$ is output, the navigation episode is considered successful. 

Zero-shot object navigation task divides the target objects into a training set $P_{train} = \{p_1, p_2, \cdots, p_n\}$ and a test set $P_{test} = \{p_{n+1}, p_{n+2}, \cdots, p_{n+m}\}$. The objects in the test set are only available during the testing process.

\subsection{Thinking Inputs}
\label{sec:Thinking Inputs}

Each thinking’s input is the most important inductive bias for the corresponding meta-ability. Input features should be as concise as possible while meeting the requirements of the meta-abilities. As shown in the thinking boxes in Figure~\ref{architecture}, we select five specialized inputs from the agent’s available information based on the characteristics of the meta-abilities.

\begin{enumerate}[1)]\setlength{\itemsep}{-0.1cm}
\item \textbf{Intuition thinking inputs} $IT_i \in \mathbb{R}^{7\times 7\times 512}$ are extracted from the first-person perspective image using a fixed-weight ResNet18 \cite{ResNet18}.

\item \textbf{Search thinking inputs} $ST_i \in \mathbb{R}^{N\times 262}$ are the object visual and position features extracted from the image using DETR \cite{DETR}. 

\item \textbf{Navigation thinking inputs} $NT_i \in \mathbb{R}^{D_{n}\times 9}$ follow the target-oriented memory graph (TOMG) proposed in \cite{DAT}. Navigation thinking only focuses target-related information; thus the TOMG is composed of the target bounding box and the agent’s coordinates on the visited target-visible nodes.

\item \textbf{Exploration thinking inputs} $ET_i \in \mathbb{R}^{D_{e}\times 4}$ are the agent's historical positions and camera angles.

\item \textbf{Obstacle thinking inputs} $OT_i \in \mathbb{R}^{D_{o}\times 2}$ are the positions of known unreachable nodes. When the agent attempts to reach a certain node and fails, it will record that node as unreachable.
\end{enumerate}

$N$ is the number of objects. $D_n$, $D_e$, $D_o$ respectively represent the number of visited target-visible nodes, visited nodes and known unreachable nodes.

\subsection{Thinking Embedding}
\label{sec:Thinking Embedding}
Thinking embedding abstracts thinking inputs into the semantic space for decision making.
Past works \cite{CKR, DUET} have introduced various prior knowledge into thinking encoding networks to guide models' attention. However, our MT method only uses a minimal amount of encoding techniques based on the characteristics of each meta-ability, highlighting the advantage of the MAD paradigm itself.

\paragraph{Intuition Thinking}
A simple learnable pointwise convolution directly encodes the input ResNet features:
\begin{equation}
    IT_o = \delta (Conv(IT_i))
\end{equation}
where $Conv$ refers to the pointwise convolution and $\delta$ represents the ReLU nonlinearity \cite{ReLU}.

\paragraph{Search Thinking}
Search thinking aims to enable the agent to quickly capture the target with the fewest steps when the target is not in view. In order to have the object association ability, we adopt the unbiased directed object attention (DOA) graph $G_t \in \mathbb{R}^{N\times N}$ proposed in \cite{DOA} to assign weights to each object. We extract the object's attention weight vector $G_t^p$ from $G_t$ based on the target $p$, and assign it to each encoded object feature:
\begin{equation}
    ST_o = \delta (ST_i W^{ST})\odot G_{t}^{p}
\end{equation}
$W^{ST}$ is a learnable parameter matrix and $\odot$ allows each object feature to be multiplied by its corresponding attention coefficient.

\paragraph{Navigation Thinking}
Navigation thinking requires the ability to memorize, locate and navigate to the target. We borrow the target-aware multi-scale aggregator (TAMSA) proposed in \cite{DAT} to map the target observation information of different positions into the target orientation relative to the current state.

First, the decisions (e.g. rotate right) are made relative to the current agent’s state $(x_c, y_c, \theta _c , \beta _c)$, so we self-center the agent's states $(x_i, y_i, \theta _i , \beta _i)$ stored in TOMG.
\begin{equation}
\begin{aligned}
(\widetilde {x}_i, \widetilde {y}_i)&=(x_i,y_i)-(x_c,y_c)\\
(\widetilde {\theta} _i^{x}, \widetilde {\beta } _i^{x})&=sin((\theta _i,\beta _i)-(\theta  _c,\beta _c))\\
(\widetilde {\theta} _i^{y}, \widetilde {\beta } _i^{y})&=cos((\theta _i,\beta _i)-(\theta _c,\beta _c))\;\;\;i\in \Delta _M
\end{aligned}
\end{equation}
where $\Delta _M$ represents the index collection of target-visible nodes. To ensure that the angle and position coordinates have the same order of magnitude, we use $sin$ and $cos$ to normalize the angle coordinates to $[-1, 1]$. After this egocentric coordinate transformation, we obtain egocentric TOMG features $ \widetilde{NT_i} \in \mathbb{R}^{D_{n}\times 11}$. Similarly, the agent states in exploration thinking and obstacle thinking also need egocentric transformation as described above.

The subsequent encoding process is represented as:
\begin{gather}
NT_o = H^{T}F_{NT}(\widetilde{NT_i})\odot F_{E}(E) \\
H = \sum_{j=1}^{3}TCN_j(\widetilde{NT_i})
\end{gather}
$H \in \mathbb{R}^{D_{n}\times 1}$ is obtained by summing different scale kernels that are generated by using the multi-scale temporal convolution networks (TCNs) to process $\widetilde{NT_i}$.
$F_{NT}(\cdot)$ maps $\widetilde{NT_i}$ to a higher-dimensional feature space. Since navigation thinking needs to adaptively change when searching for different targets, the one-hot target index $E$ is encoded by two fully connected (FC) layers $F_{E}(\cdot )$ to generate a channel-wise activation vector which recalibrates channel-wise feature responses.

\paragraph{Exploration Thinking}

We hope that the agent can use exploration thinking to more efficiently explore the environment and avoid repeated exploration. After self-centering the agent's state, we also introduce two inductive biases by polarizing the coordinates. (\romannumeral1) Through the distance of polar coordinates, the network could learn that historical states closer to the current node are more important. (\romannumeral2) Through the angle of polar coordinates, the network could learn that traveling in the direction of less exploration can gain more knowledge of the scene. Subsequently, we use two FC layers $F_{ET}(\cdot)$ and a global average pooling layer to obtain the output of exploration thinking.
\begin{equation}
ET_o=\frac{1}{D_e} \sum_{l=1}^{D_e}F_{ET}(g(ET_i\left \langle l \right \rangle))
\end{equation}
where $\left \langle l \right \rangle$ retrieves the feature of the $l$-th node from the historical memory graph, and $g(\cdot)$ represents the process of converting a cartesian coordinate system to a polar coordinate system.

\paragraph{Obstacle Thinking}
Previous approaches commonly suffered from the issue of repeatedly colliding with the same obstacle, leading to deadlock. Our obstacle thinking helps the agent quickly escape from deadlock states by memorizing collided obstacles. The overall encoding process is similar to that of exploration thinking. 
\begin{equation}
OT_o=\frac{1}{D_o} \sum_{l=1}^{D_o}F_{OT}(g(OT_i\left \langle l \right \rangle))
\end{equation}

Dropout layer is added after the more complex intuition thinking, search thinking, and navigation thinking in the above five thinking encoding networks.

\begin{figure}[t]
\begin{center}
\centerline{\includegraphics[width=0.85\columnwidth]{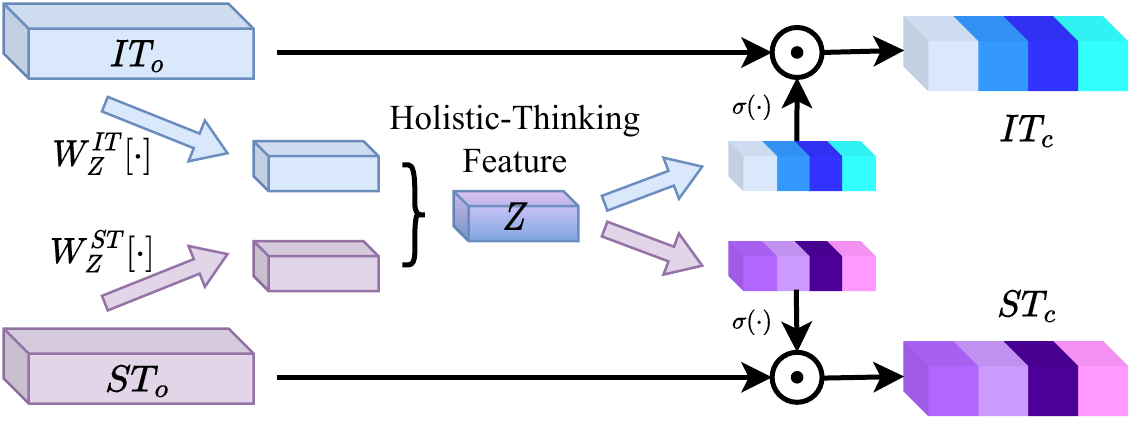}}
\caption{\textbf{Multiple thinking collaboration (MTC) module between two thinking.}  The increase in the number of thinking does not affect the overall structure. We first extract the holistic-thinking feature from the outputs of multiple thinking. Then, the channel activation vector is generated for each thinking, which recalibrates the thinking features.}
\label{MTC}
\end{center}
\vskip -0.2in
\end{figure}

\subsection{Multiple Thinking Collaboration (MTC)}
\label{sec:MTC}
Although we decouple the meta-abilities required for object navigation, cooperation between the meta-ability thinking is still necessary. For instance, when the search thinking discovers that the target is to the right of the agent, the obstacle thinking needs to give the obstacles on the right more attention.
Therefore, we design a multiple thinking collaboration (MTC) module (Figure~\ref{MTC}) to transmit shared information between different thinking.

The MTC module primarily recalibrates the channel weights of each thinking using the condensed information from all thinking. Initially, we squeeze the outputs of multiple thinking into a holistic-thinking feature representation $Z$:
\begin{equation}
Z = \delta (W_Z\left [ IT_o, ST_o, NT_o, ET_o, OT_o \right ] +b_Z)
\end{equation}
Then, excitation signals for each thinking are generated to recalibrates each thinking output $\mathcal{X}T_o$:
\begin{equation}
\mathcal{X}T_c = \mathcal{X}T_o \odot \sigma (W_{\mathcal{X}T}Z+b_{\mathcal{X}T}) \;\;\; \mathcal{X}\rightarrow I,S,N,E,O 
\end{equation}
where $\sigma(\cdot)$ represents the sigmoid activation function.

\subsection{Policy Learning}
After holistic-thinking recalibration, all thinking features need to be integrated into a unified representation vector:
\begin{equation}
G=\delta (W_G[IT_c, ST_c, NT_c, ET_c, OT_c]_{LN}+b_G)
\end{equation}
where $[\cdot]_{LN}$ concatenates the final output features of each thinking and uses layer normalization to stabilize the forward input distribution and backpropagation gradient. Finally, the multiple thinking joint representation $G$ is used to learn an LSTM \cite{LSTM} action policy $\pi(a_t|G_t, p)$. Following the previous works \cite{SP, DOA}, we treat this task as a reinforcement learning problem and utilize the asynchronous advantage actor-critic (A3C) algorithm \cite{A3C}.

\subsection{Meta-Ability Reward}
\label{sec:Meta-Ability Reward}
Our model is supervised by two types of reward: base reward $R_B$ and meta-ability reward $R_{MA}$. Similar to the previous work \cite{HOZ}, $R_B$ is composed of three parts. (\romannumeral1) We penalize each step with a small negative reward -0.01. (\romannumeral2) To encourage movement, if the agent outputs $MoveAhead$, a positive reward of 0.01 is given. (\romannumeral3) If any object instance from the target object category is reached within a certain number of steps, the agent receives a large positive reward 5.0. Meta-ability reward is designed for the goals that each meta-ability needs to achieve. 
\begin{equation}
R_{MA} = R_s + R_n + R_e + R_o
\end{equation}
\paragraph{Search Reward $R_s$}
If the target object is correctly identified in the field of view, $R_s=0.01$, otherwise $R_s=0$.
\paragraph{Navigation Reward $R_n$}
Once the target has been located, if the chosen action allows the agent to move closer to the target, $R_n=0.01$, otherwise $R_n=0$.
\paragraph{Exploration Reward $R_e$}
If the agent repeatedly reaches the same state, $R_e=-0.01$, otherwise $R_e=0$.
\paragraph{Obstacle Reward $R_o$}
If the agent collides with an obstacle, $R_o=-0.01$, otherwise $R_o=0$.

$R_{MA}$ enables each meta-ability thinking to more quickly capture the direction of learning. Accordingly, during the initial $C$ training episodes when guiding the model’s learning direction, the model is supervised by both meta-ability reward $R_{MA}$ and base reward $R_B$. Afterwards, when better task performance metrics are desired, the model receives supervision solely from the base reward $R_B$.

\section{Experiment}
\subsection{Experimental Setup}
\paragraph{Datasets}
AI2-Thor \cite{AI2-THOR} and RoboTHOR \cite{RoboTHOR} are our primary experimental platforms. AI2-Thor includes 30 different floorplans for each of 4 room layouts: kitchen, living room, bedroom, and bathroom. For each scene type, we use 20 rooms for training, 5 rooms for validation, and 5 rooms for testing. RoboTHOR consists of a set of 89 apartments, 75 of which are accessible. we use 60 for training and 15 for validation. RoboTHOR is a more complex version of the AI2-Thor environment, with 2.4 times larger floor area and 5.5 times longer path length. 

For zero-shot object navigation, we re-split the widely used 22 target classes \cite{MJO, COS_ZSON} into 18/4 seen/unseen and 14/8 seen/unseen classes. We train the model with seen object classes as the targets and test the model with unseen object classes as the targets. 

\paragraph{Evaluation Metrics}
We use the success rate (SR), success weighted by path length (SPL) \cite{SPL} metrics to evaluate our method. SR indicates the success rate of the agent in completing the task, which is formulated as $SR = \frac{1}{F}\sum_{i=1}^{F}Suc_i$, where $F$ is the number of episodes and $Suc_i$ indicates whether the $i$-th episode succeeds. SPL considers the path length more comprehensively and is defined as $SPL = \frac{1}{F}\sum_{i=1}^{F}Suc_{i}\frac{L_i^*}{max(L_i, L_i^*)}$, where $L_i$ is the path length taken by the agent and $L_i^*$ is the theoretical shortest path. 

\paragraph{Implementation Details}
The model with only intuition thinking 
(IT) and base reward $R_B$ is our baseline. We train our model with 18 workers on 2 RTX 2080Ti Nvidia GPUs, in a total of 3M navigation episodes. The dropout rate is set to 0.3, and the meta-ability reward $R_{MA}$ is only utilized in the first 0.2M ($C$) episodes. We report the results for all targets (ALL) and for a subset of targets ($L\ge 5$) with optimal trajectory lengths greater than 5.


\begin{table}[t]
\centering
\small
\setlength\tabcolsep{5pt}
\caption{Ablation experiments for each meta-ability. Removing a meta-ability means removing the corresponding thinking and reward for the meta-ability.}
\label{tab:ablation}
\begin{tabular}{ccccc|cc|cc}
\hline
\multirow{2}{*}{IT} & \multirow{2}{*}{ST} & \multirow{2}{*}{NT} & \multirow{2}{*}{ET} & \multirow{2}{*}{OT} & \multicolumn{2}{c|}{ALL (\%)} & \multicolumn{2}{c}{$L\ge 5$ (\%)} \\
 &  &  &  &  & SR$\uparrow$ & SPL$\uparrow$ & SR$\uparrow$ & SPL$\uparrow$ \\ \hline \hline
\checkmark &  &  &  &  & 43.48 & 18.91 & 30.36 & 14.34 \\
\checkmark & \checkmark &  &  &  & 75.94 & 42.77 & 68.24 & 43.28 \\
\checkmark & \checkmark & \checkmark &  &  & 79.62 & 46.13 & 72.96 & 45.93 \\
\checkmark & \checkmark & \checkmark & \checkmark &  & 81.97 & 48.75 & 75.54 & 48.95 \\ \hline
\checkmark & \checkmark & \checkmark & \checkmark & \checkmark & \textbf{83.14} & \textbf{50.23} & \textbf{77.03} & \textbf{50.88} \\ \hline
\end{tabular}
\end{table}

\begin{table}[t]
\small
\centering
\setlength\tabcolsep{3.7pt}
\caption{Ablation experiments for the multiple thinking collaboration (MTC) module and meta-ability reward.}
\label{tab:MTC_MAR}
\begin{tabular}{c|l|cc|cc}
\hline
\multirow{2}{*}{ID} & \multirow{2}{*}{Method} & \multicolumn{2}{c|}{ALL (\%)} & \multicolumn{2}{c}{$L\ge 5$ (\%)} \\
 &  & SR$\uparrow$ & SPL$\uparrow$ & SR$\uparrow$ & SPL$\uparrow$ \\ \hline \hline
1 & Complete MT & 83.14 & 50.23 & 77.03 & 50.88 \\ \hline
2 & MT $\rightarrow$ No MTC & 82.26 & 50.14 & 76.17 & 50.52 \\
3 & MT $\rightarrow$ No $R_{MA}$ & 81.96 & 49.54 & 76.22 & 49.93 \\
4 & $R_B+R_{MA}$ (All Episodes) & 81.31 & 48.29 & 75.36 & 48.44 \\ \hline
\end{tabular}
\end{table}

\subsection{Ablation Experiments}
\paragraph{Meta-Ability Ablation}
The object navigation task is decomposed into a total of five meta-abilities, which are ablated in Table~\ref{tab:ablation}. Based on the MAD structure, search ability is the most important meta-ability, followed by navigation ability, with exploration ability and obstacle ability playing a supportive role. 

\paragraph{MTC and Meta-Ability Reward Ablation}
Table~\ref{tab:MTC_MAR} shows the ablation results where the MTC module and the meta-ability reward $R_{MA}$ are removed in the second and third rows, respectively. We observe that the MTC module has a greater impact on SR, and the meta-ability reward improves both SR and SPL. The results of the fourth row in Table~\ref{tab:MTC_MAR}, in which both base reward $R_B$ and meta-ability reward $R_{MA}$ are used throughout the entire training process, suggest that using meta-ability reward in the later stages of training can divert the model's pursuit of the final goal (finding the object via the shortest path) and result in a disconnection between the reward and actual performance.

\subsection{Comparative Analysis of Different Targets}
\begin{figure}[t]
\vskip 0.2in
\begin{center}
\centerline{\includegraphics[width=\columnwidth]{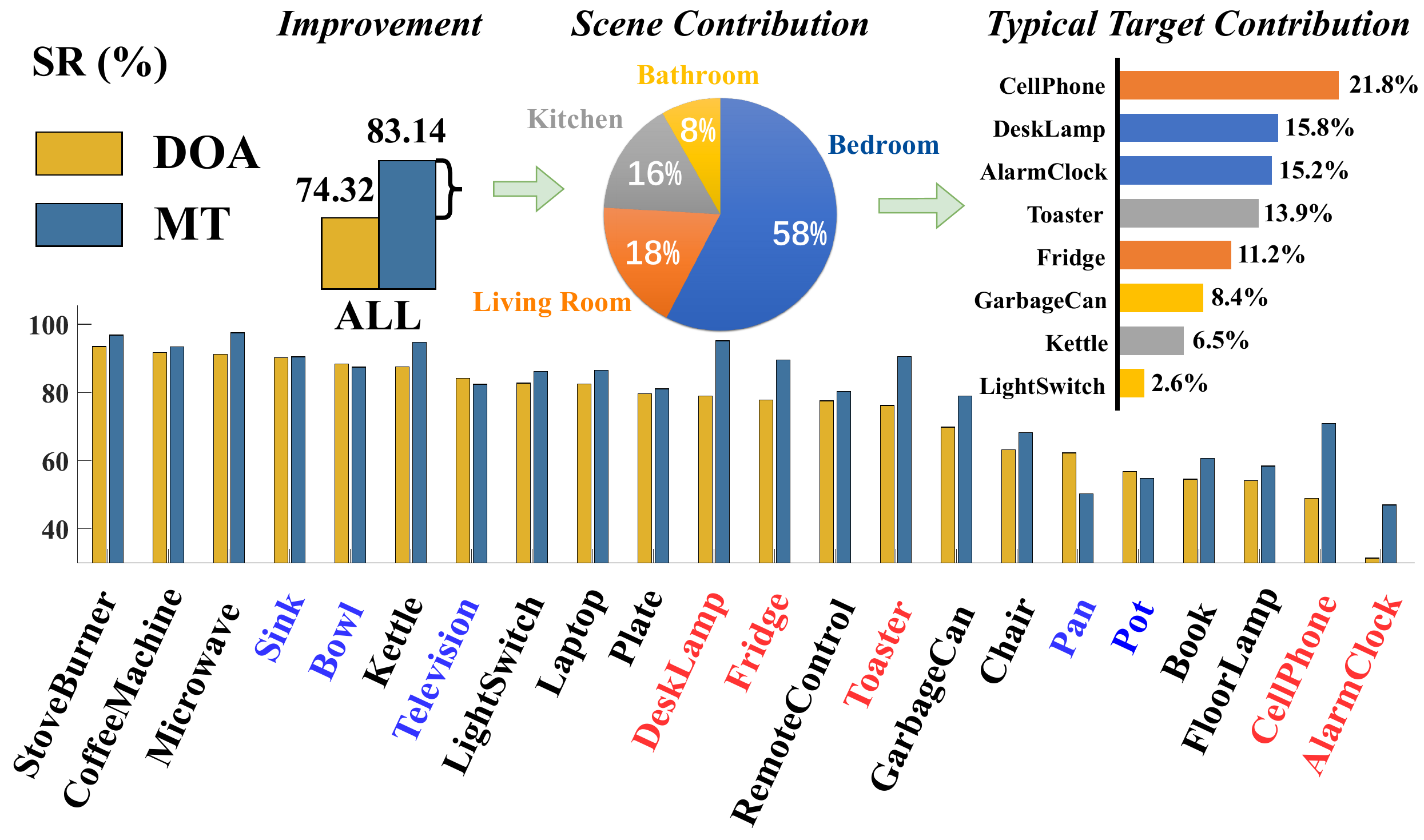}}
\caption{\textbf{Comparison of our MT method with the DOA method in terms of SR index for each individual target.} 
The red and blue markers indicate the targets with highest and lowest performance improvement of the MT method respectively. The pie chart shows the contribution of each scene to overall SR improvement. Subsequently, two objects with highest contribution from each scene are plotted in a bar chart.}
\label{each_object_sr}
\end{center}
\vskip -0.2in
\end{figure}

Figure~\ref{each_object_sr} compares the SR of our MT method and the current SOTA method (DOA \cite{DOA}) for each target object. Previous methods perform poorly for small objects and objects in complex environments (e.g. bedroom). The five target objects (labeled in red) that benefit most from our method are mostly previously unresolved targets. The five target objects (labeled in blue) that benefit least from our method are mostly common in the simpler kitchen scene. It is observable from the pie chart that our MT model makes a much greater overall contribution to SR improvement in complex scenes (e.g. bedroom, living room) compared to simple scenes (e.g. kitchen, bathroom). These findings indicate that our MT method can address multifaceted decision-making challenges in complex environments via flexible meta-abilities. More experiments are in Appendix~\ref{sec:D} and~\ref{sec:Appendix:E}.

\begin{table}[t]
\centering
\scriptsize
\setlength\tabcolsep{3pt}
\caption{Comparison with target-specific SOTA methods on the AI2-Thor / RoboTHOR  datasets.}
\label{tab:SOAT_standard}
\begin{tabular}{c|c|cc|cc|c}
\hline
\multirow{2}{*}{ID} & \multirow{2}{*}{Method} & \multicolumn{2}{c|}{ALL (\%)} & \multicolumn{2}{c|}{$L\ge 5$ (\%)} & \multirow{2}{*}{\begin{tabular}[c]{@{}c@{}}Episode \\ Time (s)$\downarrow$\end{tabular}}  \\
 &  & SR$\uparrow$ & SPL$\uparrow$ & SR$\uparrow$ & SPL$\uparrow$ &   \\ 
 \hline \hline
\multirow{2}{*}{\uppercase\expandafter{\romannumeral1}} & SSCNav & 77.14/38.12 & 31.09/14.10 & 71.73/33.46 & 34.33/11.04 & 1.34/4.14  \\
 & PONI & 78.58/38.42 & 33.78/16.30 & 72.92/34.72 & 36.40/13.22 & 1.59/4.58  \\ \hline
\multirow{2}{*}{\uppercase\expandafter{\romannumeral2}} & OMT & 71.13/32.17 & 37.27/20.09 & 61.94/25.33 & 38.19/18.16 & 0.64/2.01  \\
 & VGM & 73.95/35.82 & 40.69/23.71 & 64.07/27.22 & 40.73/19.54 & 0.73/2.46  \\ \hline
\uppercase\expandafter{\romannumeral3} & TPN & 67.32/30.51 & 37.01/18.62 & 58.13/23.89 & 35.90/14.91 & 0.24/0.77  \\ 
\hline
\multirow{3}{*}{\uppercase\expandafter{\romannumeral4}} & HOZ & 68.53/31.67 & 37.50/19.02 & 60.27/24.32 & 36.61/14.81 & 0.28/0.81  \\
 & VTNet & 72.24/33.92 & 44.57/23.88 & 63.19/26.77 & 43.84/19.80 & 0.32/1.33  \\
 & DOA & 74.32/36.22 & 40.27/22.12 & 67.88/30.16 & 40.36/18.32 & 0.33/1.25  \\ \hline
\rowcolor{gray!30}
\uppercase\expandafter{\romannumeral5} & \textbf{MT} & \textbf{83.14/42.80} & \textbf{50.23/29.07} & \textbf{77.03/37.85} & \textbf{50.88/23.16} & 0.35/1.20  \\ 
\hline
\end{tabular}
\end{table}

\begin{table}[t]
\small
\caption{Comparison with target-agnostic 
zero-shot SOTA methods on the AI2-Thor datasets. }
\label{tab:SOTA_ZSON}
\begin{tabular}{c|c|cccc}
\hline
\multirow{3}{*}{Method} & \multirow{3}{*}{\begin{tabular}[c]{@{}c@{}}Seen/Unseen\\ split\end{tabular}} & \multicolumn{4}{c}{Unseen Classes} \\ \cline{3-6} 
 &  & \multicolumn{2}{c|}{ALL (\%)} & \multicolumn{2}{c}{$L\ge 5$ (\%)} \\
 &  & SR$\uparrow$ & \multicolumn{1}{c|}{SPL$\uparrow$} & SR$\uparrow$ & SPL$\uparrow$ \\ \hline \hline
Random & 18/4 & 9.76 & \multicolumn{1}{c|}{2.03} & 0.82 & 0.27 \\
ZER & 18/4 & 31.28 & \multicolumn{1}{c|}{15.06} & 25.74 & 14.60 \\
ZSON & 18/4 & 57.32 & \multicolumn{1}{c|}{20.94} & 46.43 & 21.78 \\ \hline
\rowcolor{gray!30}
\textbf{MT-ZS} & 18/4 & \textbf{68.61} & \multicolumn{1}{c|}{\textbf{27.95}} & \textbf{57.53} & \textbf{29.28} \\ \hline
Random & 14/8 & 7.70 & \multicolumn{1}{c|}{3.19} & 0.44 & 0.08 \\
ZER & 14/8 & 24.62 & \multicolumn{1}{c|}{10.42} & 14.33 & 8.99  \\
ZSON & 14/8 &  52.74 &  \multicolumn{1}{c|}{18.11} &  33.53 & 14.38  \\ \hline
\rowcolor{gray!30}
\textbf{MT-ZS} & 14/8 & \textbf{62.40} & \multicolumn{1}{c|}{\textbf{24.08}} &  \textbf{46.58} & \textbf{23.76} \\ \hline
\end{tabular}
\end{table}

\subsection{Comparisons with the State-of-the-Art}

\paragraph{Target-Specific Typical Object Navigation}
In Table~\ref{tab:SOAT_standard}, our MT model is compared with the four categories of SOTA models. \textbf{(\uppercase\expandafter{\romannumeral1}) SLAM methods.} The real-time construction of a semantic map and sub-goal path planning technique enhance the interpretability of these methods. However, due to the significant cost of exploring the environment, the time required for each episode is several times that of other methods. \textbf{(\uppercase\expandafter{\romannumeral2}) Memory methods.} The explicit memory of long-term historical information enhances the model's exploration and navigation ability. Despite this, existing memory methods have a large amount of redundancy, resulting in poor generalization. 
\textbf{(\uppercase\expandafter{\romannumeral3}) Deadlock-specialized methods.} Deadlock states occur frequently while navigating. Although incorporating a deadlock-specialized module resolves some issues, it disrupts the overall cohesiveness of model training. 
\textbf{(\uppercase\expandafter{\romannumeral4}) Association methods.} Object or region association is the most common inductive bias. Narrowly, association methods only serve to enhance search ability, with minimal effect on other abilities. 
Under the guidance of the MAD paradigm, our MT method explicitly decouples the various meta-abilities of the object navigation task, theoretically unifying the above four types of models. In comparison to the SOTA method (DOA \cite{DOA}) with similar computational complexity, our MT method brings an overall 8.82/6.58 and 9.96/6.95 improvement in SR and SPL (AI2-Thor / RoboTHOR, \%), respectively.

\paragraph{Target-Agnostic Zero-Shot Object Navigation}
In contrast to typical object navigation tasks, the target objects in the zero-shot task are not visible during training. Consequently, we replace the object one-hot encoding in the MT model with the cosine similarity of glove embedding to the target \cite{COS_ZSON}, resulting in the MT-ZS model. In comparison with the CLIP-based ZER model \cite{ZER} and the class-unrelated ZSON model \cite{COS_ZSON}, our MT-ZS model based on the MAD paradigm exhibits clear advantages (Table~\ref{tab:SOTA_ZSON}). The success of our MAD paradigm in the zero-shot task demonstrates its effectiveness for embodied AI tasks with high generalization difficulty as well. More experiments are in Appendix~\ref{sec:C}.

\begin{figure*}[t]
\vskip 0.2in
\begin{center}
\centerline{\includegraphics[width=\textwidth]{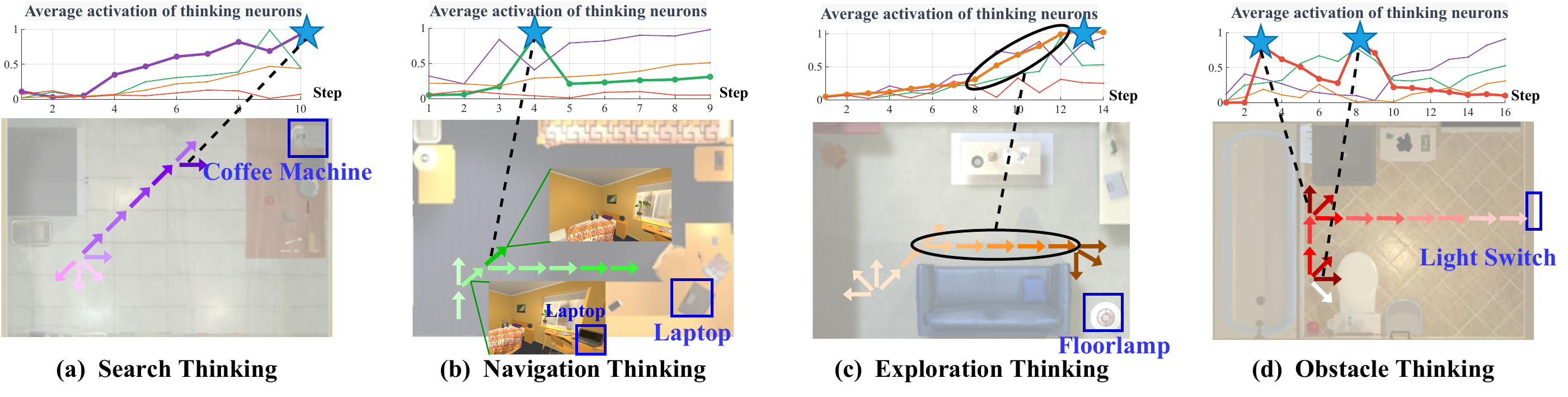}}
\caption{\textbf{We visualize the average activation of each thinking’s neurons during navigation.} The depth of the arrow color represents the average value of the current thinking’s output neurons, corresponding to the line chart above. The blue pentagram signifies the step in the path where thinking is most active.}
\label{visual_ai2thor}
\end{center}
\vskip -0.2in
\end{figure*}

\subsection{Meta-ability Qualitative Analysis}
\label{sec:Meta-ability Qualitative Analysis}
Our MT model makes decisions based on the synthesis of various meta-abilities during navigation. In Figure~\ref{visual_ai2thor}, we visualize the mean value of the thinking output neurons $\mathcal{X}T_c$ at each step to explore how each thinking influences model inference in different scenarios.  Intuition thinking exhibits no discernible pattern, so in this case we only illustrate the other four thinking. 
(a) Search thinking is activated when the target or target-related objects are observed, and becomes increasingly active as the target is approached.
(b) When the target object is suddenly lost from the agent's field of view, the level of navigation thinking becomes greatly heightened, enabling the agent to quickly reacquire the target object.
(c) Continuous forward motion quickly activates exploration thinking. 
(d) Obstacle thinking is maximally activated when the agent encounters an obstacle, and the level of activation gradually decreases as the distance from the obstacle increases.
Each thinking’s excitation provides a clearer explanation for the model's decision-making process based on meta-abilities. More analysis is in Appendix~\ref{Appendix:F}.

\begin{table}[t]
\scriptsize
\centering
\setlength\tabcolsep{4.5pt}
\caption{Quantitative comparison of meta-abilities across different models.}
\label{tab:meta-ability-metics}
\begin{tabular}{c|cccc}
\hline
\multirow{2}{*}{Method} & \multicolumn{4}{c}{ALL (\%)} \\
 & SSR (S)$\uparrow$ & NSNLP (N)$\uparrow$ & REP (E)$\downarrow$ & CP (O)$\downarrow$ \\ \hline
Baseline & $91.35$ & $23.15$ & $5.28$ & $12.74$ \\
DOA & $95.82_{(+4.47)}$ & $44.11_{(+20.96)}$ & $7.14_{(+1.86)}$ & $10.26_{(-2.48)}$ \\
\rowcolor{gray!30}
MT & $97.76_{(+6.14)}$ & $51.39_{(+28.24)}$ & $4.03_{(-1.25)}$ & $4.93_{(-7.81)}$ \\ \hline
\end{tabular}
\end{table}
\subsection{Meta-Ability Quantitative Analysis}
\paragraph{Meta-Ability Metrics}
In order to quantitatively evaluate the meta-abilities of each model, we define four meta-ability metrics: (\romannumeral1) search success rate (SSR): the success rate in finding the target; (\romannumeral2) navigation success weighted by navigation path length (NSNPL): SPL during the navigation phase after finding the target; (\romannumeral3) repeated exploration probability (REP): probability of reaching the same state repeatedly; (\romannumeral4) collision probability (CP): proportion of actions resulting in collision with obstacles. Larger SSR and NSNPL values indicate stronger search and navigation abilities, while smaller REP and CP values indicate stronger exploration and obstacle abilities. More detailed explanations are in Appendix~\ref{sec:B}.

\paragraph{Analysis}
As shown in Table~\ref{tab:meta-ability-metics}, our MT model performs significantly better in each meta-ability than the other models. Current SOTA method DOA primarily utilizes object association to enhance search ability, however, REP indicates that the exploration ability of the DOA method has decreased relative to the baseline model. This phenomenon suggests that without decoupling meta-abilities, enhancing one meta-ability may lead to weakening other meta-abilities. It is noteworthy that our MT model only employs a small fraction of the inductive bias used in the DOA model to enhance search ability (Sec.~\ref{sec:Thinking Embedding}), yet the search ability of the MT model outperforms that of the DOA model. This finding leads us to believe that the thinking specificity promoted by the MAD paradigm can amplify the impact of each inductive bias on the corresponding meta-ability.

\section{Limitation}
There are still some limitations to this paper. (\romannumeral1) The selection of meta-abilities depends on human experience, so how to decouple the more abstract meta-abilities is still an open problem.  (\romannumeral2) MAD is only applied and experimented on the object navigation task in this paper, and we expect that researchers can expand it to more embodied AI tasks. (\romannumeral3) How meta-ability thinking affects the model’s decision-making still has many directions worthy of exploration. 

\section{Conclusion}

This paper proposes the meta-ability decoupling (MAD) paradigm, which guides researchers to design and analyze object navigation models from the perspective of meta-abilities. Based on MAD, we design the multiple thinking (MT) model, which significantly outperforms SOTA methods in both typical and zero-shot object navigation tasks. 
Additionally, we conduct a qualitative and quantitative interpretability analysis of the MT model at the meta-ability level.
Beyond the object navigation, the underlying principles are theoretically generalizable to other embodied AI tasks.

\nocite{langley00}

\bibliography{example_paper}
\bibliographystyle{icml2022}

\newpage
\appendix
\onecolumn
\section{Related Works}
\label{sec:A}
In the main text, typical object navigation methods are classified into four categories, and the problems addressed by these four categories are summarized. In this chapter, we will provide a more detailed introduction to the representative models within each category. 

\subsection{Association Methods}
Association methods can be divided into three categories, object association, zone association, and room association, from detailed to rough. Representative object association methods include SP \cite{SP}, DOA \cite{DOA}, and CKR \cite{CKR}.  SP and DOA only rely on data in the environmental scene to model the spatial correlation between known objects. CKR incorporates semantic correlation between objects from a large-scale external knowledge graph into the model. HOZ \cite{HOZ} proposes the zone association to guide an agent in a coarse-to-fine manner. BRM \cite{BRM} takes the form of a probabilistic room relation graph to capture the layout prior. 

\subsection{Memory Methods}
Memory methods explicitly store a large amount of historical information, such as visual features, coordinate features, object features, etc. VGM \cite{VGM} is constructed incrementally based on the similarities among the unsupervised representations of observed images, and these representations are learned from an unlabeled image dataset. OMT \cite{OMT} uses transformer to salient objects stored in memory. DUET \cite{DUET} proposes a joint long-term action planning to enable efficient exploration in global action space. 

\subsection{SLAM Methods}
Traditional navigation methods in known environments are all dependent on SLAM maps, thus exploring unknown environments through real-time mapping is also a viable approach. Due to the high cost of mapping, most methods now choose more rough semantic maps. GOSE \cite{GOSE} builds an episodic semantic map and uses it to explore the environment efficiently based on the goal object category. SSCNav \cite{SSCNav} explicitly models scene priors using a confidence-aware semantic scene completion module to complete the scene and guide the agent’s navigation planning. PONI \cite{PONI} proposes a network that predicts two complementary potential functions conditioned on a semantic map and uses them to decide where to look for an unseen object. 

\subsection{Deadlock-Specialized Methods}
Deadlock-specialized modules are frequently a part of the overall method to assist the agent in breaking out cyclic states. TPN \cite{TPN} employs a pre-trained primary model to explore the environment and provides expert actions for deadlock states. SAVN \cite{SAVN} uses the similarity of observation data as the basis for determining the success of actions and incorporates it into the loss function. 

\section{Meta-Ability Metrics}
\label{sec:B}
In the main text, we introduce new metrics to evaluate four meta-abilities. In this chapter, we will provide a more detailed explanation of these four metrics. In order to differentiate search ability and navigation ability, we divide the entire episode into "search for" phase and "navigate to" phase based on the first target-visible frame as the boundary. The agent primarily relies on its search ability to locate the target object during the "search for" phase. Once the target object is observed, the agent enters the "navigation to" phase and primarily relies on its navigation ability to navigate to the location of the target object. 

\paragraph{Search Ability Metric}
SSR is the success rate for the “search for" phase and is formulated as
\begin{equation}
SSR = \frac{1}{F}\sum_{i=1}^{F}Nav_i
\end{equation}
where $Nav_i$ indicates whether the $i$-th episode enters the “navigate to" phase. 

\paragraph{Navigation Ability Metric}
NSNPL considers the navigation efficiency during the “navigate to" phase and is defined as:
\begin{equation}
NSNPL=\frac{1}{F_{Nav}}\sum_{i=1}^{F}Suc_i Nav_i\frac{L_i^{*Nav}}{max(L_i^{Nav}, L_i^{*Nav})}
\end{equation}
where $Suc_i$ indicates whether the $i$-th episode succeeds and $F_{Nav}$ is the number of episodes that enter the “navigate to" phase. $L_i^{Nav}$ is the path length in the “navigate to" phase and $L_i^{*Nav}$ is the theoretical shortest path length in the “navigate to" phase. During testing, we calculate $L_i^{*Nav}$ in real time according to the starting position of the “navigate to" phase (the position where the agent first recognizes the target) in each task path. Intuitively, NSNPL can be conceptualized as the SPL of “navigate to" phase.

\paragraph{Exploration Ability Metric}
REP, which utilizes the probability of the agent returning to previously visited states, reflects the efficiency of exploring the environment. 
\begin{equation}
REP = \frac{1}{F}\sum_{i=1}^{F}\frac{L_i - RS_i}{L_i}
\end{equation}
where $RS_i$ is the number of distinct agent states encountered in the $i$-th episode.

\paragraph{Obstacle Ability Metric}
In the real world, collisions with obstacles are to be avoided as much as possible. CP reflects the proportion of actions that resulted in collisions with obstacles throughout the entire episode.
\begin{equation}
CP = \frac{1}{F}\sum_{i=1}^{F}\frac{OA_i}{L_i}
\end{equation}
where $OA_i$ is the number of obstacle collisions that occurred in the $i$-th episode.

\section{Comparisons with the State-of-the-Art}
\label{sec:C}
\subsection{Target-Specific Typical Object Navigation}
In the main text, we only compare the SR and SPL metrics, but the analysis of meta-ability indicators for various methods is insufficient. Tables~\ref{tab:SOTA_AI2} and Tables~\ref{tab:SOTA_ROBO} respectively comprehensively present the performance metrics of various methods on the AI2-Thor and RoboTHOR datasets. 
SLAM methods (\uppercase\expandafter{\romannumeral1}) and deadlock-specialized methods (\uppercase\expandafter{\romannumeral3}) belong to modular methods, while memory methods (\uppercase\expandafter{\romannumeral2}) and association methods (\uppercase\expandafter{\romannumeral4}) belong to end-to-end methods.
Our MAD paradigm, while ensuring end-to-end training, incorporates the advantages of the above methods, and provides a clear theoretical framework for future researches. 

\paragraph{(\uppercase\expandafter{\romannumeral1}) SLAM Methods}
SLAM methods based on the AI2-Thor and RoboTHOR platforms are relatively rare, therefore, we adapt the SOTA methods (SSCNav \cite{SSCNav} and PONI \cite{PONI}) from the Habitat \cite{Habitat} platform to the AI2-Thor and RoboTHOR datasets. SLAM methods commonly use the form of waypoint prediction to guide the agent's navigation. This discrete navigation mode greatly prolongs the path to search for the target, thus reducing the overall SPL. More seriously, building an accurate map requires a significant amount of computation resources and exploration time, resulting in several times longer episode time compared to other methods. However, it is clear that SLAM methods obtain a strong navigation ability (NSNPL). Because once SLAM methods correctly establish a semantic map of the target and its surroundings, navigating to the target location becomes much easier. Another important reason why SLAM methods are favored by some researchers is their strong interpretability. We hope that on the basis of the MAD paradigm, the interpretability of end-to-end methods in navigation tasks will be gradually improved. 

\paragraph{(\uppercase\expandafter{\romannumeral2}) Memory Methods}
Currently, most memory methods are a crude form of modeling historical memory. Although mining meta abilities from all available historical information may enhance the overall ability of the model, particularly in terms of exploration ability (REP), the redundant information structure decreases generalizability and even affects other meta-abilities. The exploration thinking in our MT model draws on the historical memory structure in memory methods, but our memory features are more streamlined, thereby reducing the learning burden of the model.

\paragraph{(\uppercase\expandafter{\romannumeral3}) Deadlock-Specialized Methods}
In \cite{SAVN}, the deadlock problem in the navigation process began to be noticed. TPN \cite{TPN} utilizes a supervised-trained deadlock escape module to make REP and CP reach 5.83/9.76 and 5.22/10.47 (AI2-Thor/RoboTHOR, \%) respectively. However, this extrinsic deadlock-specialized module requires a significant amount of human-annotated escape actions. Therefore, if dataset migration occurs, a significant annotation cost would have to be incurred again. Our MT method, while ensuring model scalability, yields REP and CP metrics that are 1.80/2.14 and 0.29/0.56 (AI2-Thor/RoboTHOR, \%) lower than the TPN method. 

\paragraph{(\uppercase\expandafter{\romannumeral4}) Association Methods}
Association methods primarily learn the intrinsic correlation between objects to accelerate the visual capture of the target object, thereby yielding a strong search ability (SSR). However, excessive focus on semantic information at the object level may overlook navigation details, as evidenced by excessively high REP and CP. Our MT method, by decoupling more meta-abilities, helps association methods optimize environment exploration and obstacle avoidance, thus improving SR and SPL by 8.82/5.95 and 9.96/7.02 (AI2-Thor/RoboTHOR, \%) respectively, with almost no additional parameters introduced. 

\begin{table*}[t]
\centering
\small
\setlength\tabcolsep{4.5pt}
\caption{Comparison with target-specific SOTA methods in AI2-Thor. \XSolidBrush indicates unacceptable resource consumption.}
\label{tab:SOTA_AI2}
\begin{tabular}{c|c|cccccc|cccccc|c}
\hline
\multirow{2}{*}{ID} & \multirow{2}{*}{Method} & \multicolumn{6}{c|}{ALL (\%)} & \multicolumn{6}{c|}{$L\ge 5$ (\%)} & \multirow{2}{*}{\begin{tabular}[c]{@{}c@{}}Episode\\ Time (s)$\downarrow$\end{tabular}} \\
 &  & SR$\uparrow$ & SPL$\uparrow$ & SSR$\uparrow$ & NSNPL$\uparrow$ & REP$\downarrow$ & CP$\downarrow$ & SR$\uparrow$ & SPL$\uparrow$ & SSR$\uparrow$ & NSNPL$\uparrow$ & REP$\downarrow$ & CP$\downarrow$ &  \\ \hline \hline
\multirow{2}{*}{\uppercase\expandafter{\romannumeral1}} & SSCNav & 77.14 & 31.09 & 89.14 & 51.72 & 5.14 & \textbf{4.58} & 71.73 & 34.33 & 89.02 & 50.73 & 7.30 & 5.41 & 1.342 \XSolidBrush \\
 & PONI & 78.58 & 33.78 & 89.48 & \textbf{52.39} & 5.29 & 4.90 & 72.92 & 36.40 & 89.13 & \textbf{51.82} & 7.64 & 5.75 & 1.591 \XSolidBrush \\ \hline
\multirow{2}{*}{\uppercase\expandafter{\romannumeral2}} & OMT & 71.13 & 37.27 & 93.17 & 41.36 & 4.62 & 9.88 & 61.94 & 38.19 & 92.23 & 42.63 & 6.81 & 10.74 & 0.645 \\
 & VGM & 73.95 & 40.69 & 93.20 & 44.21 & 4.51 & 9.30 & 64.07 & 40.73 & 92.14 & 45.97 & 6.62 & 10.14 & 0.714 \\ \hline
\uppercase\expandafter{\romannumeral3} & TPN & 67.32 & 37.01 & 91.07 & 40.24 & 5.83 & 5.22 & 58.13 & 35.90 & 90.27 & 38.69 & 8.06 & 6.89 & 0.241 \\ \hline
\multirow{3}{*}{\uppercase\expandafter{\romannumeral4}} & HOZ & 68.53 & 37.50 & 91.44 & 40.83 & 8.32 & 10.77 & 60.27 & 36.61 & 90.31 & 39.82 & 11.54 & 11.36 & 0.283 \\
 & VTNet & 72.24 & 44.57 & 94.18 & 46.74 & 7.91 & 10.71 & 63.19 & 43.84 & 92.85 & 46.15 & 10.88 & 11.52 & 0.321 \\
 & DOA & 74.32 & 40.27 & 95.82 & 44.11 & 7.14 & 10.26 & 67.88 & 40.36 & 93.92 & 44.03 & 10.39 & 10.95 & 0.334 \\ \hline
 \rowcolor{gray!30}
\uppercase\expandafter{\romannumeral5} & \textbf{MT} & \textbf{83.14} & \textbf{50.23} & \textbf{97.76} & 51.39 & \textbf{4.03} & 4.93 & \textbf{77.03} & \textbf{50.88} & \textbf{96.47} & 51.49 & \textbf{6.25} & \textbf{5.26} & 0.352 \\ \hline
\end{tabular}
\end{table*}

\begin{table*}[t]
\centering
\small
\setlength\tabcolsep{4.5pt}
\caption{Comparison with target-specific SOTA methods in RoboTHOR. \XSolidBrush indicates unacceptable resource consumption.}
\label{tab:SOTA_ROBO}
\begin{tabular}{c|c|cccccc|cccccc|c}
\hline
\multirow{2}{*}{ID} & \multirow{2}{*}{Method} & \multicolumn{6}{c|}{ALL (\%)} & \multicolumn{6}{c|}{$L\ge 5$ (\%)} & \multirow{2}{*}{\begin{tabular}[c]{@{}c@{}}Episode\\ Time (s)$\downarrow$\end{tabular}} \\
 &  & SR$\uparrow$ & SPL$\uparrow$ & SSR$\uparrow$ & NSNPL$\uparrow$ & REP$\downarrow$ & CP$\downarrow$ & SR$\uparrow$ & SPL$\uparrow$ & SSR$\uparrow$ & NSNPL$\uparrow$ & REP$\downarrow$ & CP$\downarrow$ &  \\ \hline  \hline
\multirow{2}{*}{\uppercase\expandafter{\romannumeral1}} & SSCNav & 38.12 & 14.10 & 61.37 & 35.14 & 8.93 & 10.83 & 33.46 & 11.04 & 60.91 & 33.92 & 10.17 & 13.44 & 4.145 \XSolidBrush \\
 & PONI & 38.42 & 16.30 & 58.46 & \textbf{39.83} & 8.32 & 11.22 & 34.72 & 13.22 & 58.11 & \textbf{38.44} & 11.64 & 14.27 & 4.582 \XSolidBrush \\ \hline
\multirow{2}{*}{\uppercase\expandafter{\romannumeral2}} & OMT & 32.17 & 20.09 & 61.77 & 24.51 & 7.72 & 16.45 & 25.33 & 18.16 & 57.35 & 23.82 & 9.70 & 18.83 & 2.011 \\
 & VGM & 33.95 & 22.74 & 62.10 & 25.96 & 8.21 & 15.81 & 26.82 & 19.44 & 57.51 & 24.77 & 10.66 & 18.25 & 1.984 \\ \hline
\uppercase\expandafter{\romannumeral3} & TPN & 30.51 & 18.62 & 59.64 & 20.64 & 9.76 & 10.47 & 23.89 & 14.91 & 54.64 & 19.51 & 12.28 & 13.51 & 0.769 \\ \hline
\multirow{3}{*}{\uppercase\expandafter{\romannumeral4}} & HOZ & 31.67 & 19.02 & 60.11 & 21.02 & 12.49 & 18.55 & 24.32 & 14.81 & 54.23 & 20.38 & 15.79 & 22.02 & 0.808 \\
 & VTNet & 33.92 & 23.88 & 63.29 & 28.26 & 11.26 & 17.04 & 26.77 & 19.80 & 57.72 & 27.50 & 14.63 & 21.10 & 1.325 \\
 & DOA & 36.22 & 22.12 & 64.18 & 25.88 & 11.33 & 17.14 & 30.16 & 18.32 & 61.39 & 25.11 & 14.82 & 21.52 & 1.247 \\ \hline
  \rowcolor{gray!30}
\uppercase\expandafter{\romannumeral5} & \textbf{MT} & \textbf{42.17} & \textbf{29.14} & \textbf{68.05} & 36.68 & \textbf{7.62} & \textbf{9.91} & \textbf{37.98} & \textbf{23.80} & \textbf{66.93} & 36.50 & \textbf{9.25} & \textbf{12.48} & 1.225 \\ \hline
\end{tabular}
\end{table*}

\subsection{Target-Agnostic Zero-Shot Object Navigation}
\paragraph{Task Definition}
We categorize 22 objects into two classes, namely seen and unseen. Based on the variations in classification proportion, there are two experimental setups. (\romannumeral1) 18/4: One object is extracted from each scene (bedroom, living room, kitchen and bathroom) and placed into the unseen objects category. (\romannumeral2) 14/8: Two objects are extracted from each scene and placed into the unseen objects category. Once the target set has been divided, it will not be changed. During training, seen objects are used as targets to be found, and unseen objects cannot be recognized by detectors such as target detection or instance segmentation. During testing, the agent is instructed to navigate to any given target set.

\paragraph{MT $\rightarrow$ MT-ZS}
The MT model is not suitable for the zero-shot object navigation task because it encodes object semantics using one-hot encoding and employs a fixed-size object attention matrix, thereby limiting the number of object categories from the model's perspective. The zero-shot object navigation task requires the agent to locate an arbitrary number of target objects. Therefore, we represent object semantics using Glove encoding \cite{Glove} and base object association on the semantic cosine similarity relative to the target. The continuous semantic space centered around the target allows the agent to accept requests for finding any target.

\paragraph{Results Analysis}
In the main text, we only analyze the test metrics with unseen objects as the target. In Table~\ref{tab:SOTA_ZERO}, we add the experimental results with seen objects as the target. In order to achieve zero-shot object navigation, both ZER \cite{ZER} and ZSON \cite{COS_ZSON} significantly reduce their performance on seen target sets. The MT-ZS model, which utilizes multiple meta-abilities in combination, possesses stronger navigational robustness. Therefore, our approach demonstrates a clear advantage in both searching for seen targets and unseen targets. 

\begin{table*}[t]
\small
\centering
\setlength\tabcolsep{7.7pt}
\caption{Comparing performance on seen and unseen objects with target-agnostic zero-shot SOTA methods. }
\label{tab:SOTA_ZERO}
\begin{tabular}{c|c|ccccc|ccccc}
\hline
\multirow{3}{*}{Method} & \multirow{3}{*}{\begin{tabular}[c]{@{}c@{}}Seen/Unseen\\ split\end{tabular}} & \multicolumn{5}{c|}{Unseen Classes} & \multicolumn{5}{c}{Seen Classes} \\ \cline{3-12} 
 &  & \multicolumn{2}{c|}{ALL (\%)} & \multicolumn{2}{c|}{$L \ge 5$ (\%)} & \multirow{2}{*}{\begin{tabular}[c]{@{}c@{}}Episode\\ Length$\downarrow$\end{tabular}} & \multicolumn{2}{c|}{ALL (\%)} & \multicolumn{2}{c|}{$L \ge 5$ (\%)} & \multirow{2}{*}{\begin{tabular}[c]{@{}c@{}}Episode\\ Length$\downarrow$\end{tabular}} \\
 &  & SR$\uparrow$ & \multicolumn{1}{c|}{SPL$\uparrow$} & SR$\uparrow$ & \multicolumn{1}{c|}{SPL$\uparrow$} &  & SR$\uparrow$ & \multicolumn{1}{c|}{SPL$\uparrow$} & SR$\uparrow$ & \multicolumn{1}{c|}{SPL$\uparrow$} &  \\ \hline \hline
Random & 18/4 & 9.76 & \multicolumn{1}{c|}{2.03} & 0.82 & \multicolumn{1}{c|}{0.27} & 37.523 & 9.36 & \multicolumn{1}{c|}{2.81} & 1.12 & \multicolumn{1}{c|}{0.35} & 38.146 \\
ZER & 18/4 & 31.28 & \multicolumn{1}{c|}{15.06} & 25.74 & \multicolumn{1}{c|}{14.60} & 18.492 & 25.17 & \multicolumn{1}{c|}{10.02} & 21.83 & \multicolumn{1}{c|}{10.57} & 19.447 \\
ZSON & 18/4 & 57.32 & \multicolumn{1}{c|}{20.94} & 46.43 & \multicolumn{1}{c|}{21.78} & 15.746 & 58.72 & \multicolumn{1}{c|}{18.47} & 38.44 & \multicolumn{1}{c|}{18.72} & 16.046 \\
 \rowcolor{gray!30}
\textbf{MT-ZS} & 18/4 & \textbf{68.61} & \multicolumn{1}{c|}{\textbf{27.95}} & \textbf{57.53} & \multicolumn{1}{c|}{\textbf{29.28}} & 13.215 & \textbf{66.00} & \multicolumn{1}{c|}{\textbf{27.41}} & \textbf{50.78} & \multicolumn{1}{c|}{\textbf{27.45}} & 14.742 \\ \hline
Random & 14/8 & 7.70 & \multicolumn{1}{c|}{3.19} & 0.44 & \multicolumn{1}{c|}{0.08} & 36.032 & 8.17 & \multicolumn{1}{c|}{2.94} & 0.37 & \multicolumn{1}{c|}{0.16} & 38.512 \\
ZER & 14/8 & 24.62 & \multicolumn{1}{c|}{10.42} & 14.33 & \multicolumn{1}{c|}{8.99} & 20.917 & 32.77 & \multicolumn{1}{c|}{17.25} & 30.20 & \multicolumn{1}{c|}{13.28} & 19.033 \\
ZSON & 14/8 & 52.74 & \multicolumn{1}{c|}{18.11} & 33.53 & \multicolumn{1}{c|}{14.38} & 16.680 & 59.91 & \multicolumn{1}{c|}{23.56} & 34.84 & \multicolumn{1}{c|}{20.08} & 15.892 \\
 \rowcolor{gray!30}
\textbf{MT-ZS} & 14/8 & \textbf{62.40} & \multicolumn{1}{c|}{\textbf{24.80}} & \textbf{46.58} & \multicolumn{1}{c|}{\textbf{23.76}} & 14.514 & \textbf{70.21} & \multicolumn{1}{c|}{\textbf{28.48}} & \textbf{57.63} & \multicolumn{1}{c|}{\textbf{30.46}} & 13.791 \\ \hline
\end{tabular}
\end{table*}

\section{Target Level Experiment}
\label{sec:D}
Our MT model is clearly superior to other methods in overall metrics. More detailed, we hope to understand the performance of our method under different target objects and scenario conditions. These findings not only allow us to gain a deeper understanding of our method’s advantages, but also reveal its shortcomings, providing reference for future researches.

\subsection{Experimental Setup}
In the test floorplans of AI2-Thor, we initialized 8000 tasks (scene, initial position and target object) at random. We independently count the indicators (SR, SPL, SSR, NSNPL, REP, CP) of each target object when the agent completes these 8000 tasks. To observe how the agent performs when confronted with long path tasks, we additionally extract episodes with path lengths higher than 5. The experimental results for the DOA \cite{DOA} method and our MT method are presented in Tables~\ref{tab:each_object_DOA} and Table~\ref{tab:each_object_MT}, respectively. 

\subsection{Analysis}
\begin{figure*}[t]
\vskip 0.2in
\begin{center}
\centerline{\includegraphics[width=\textwidth]{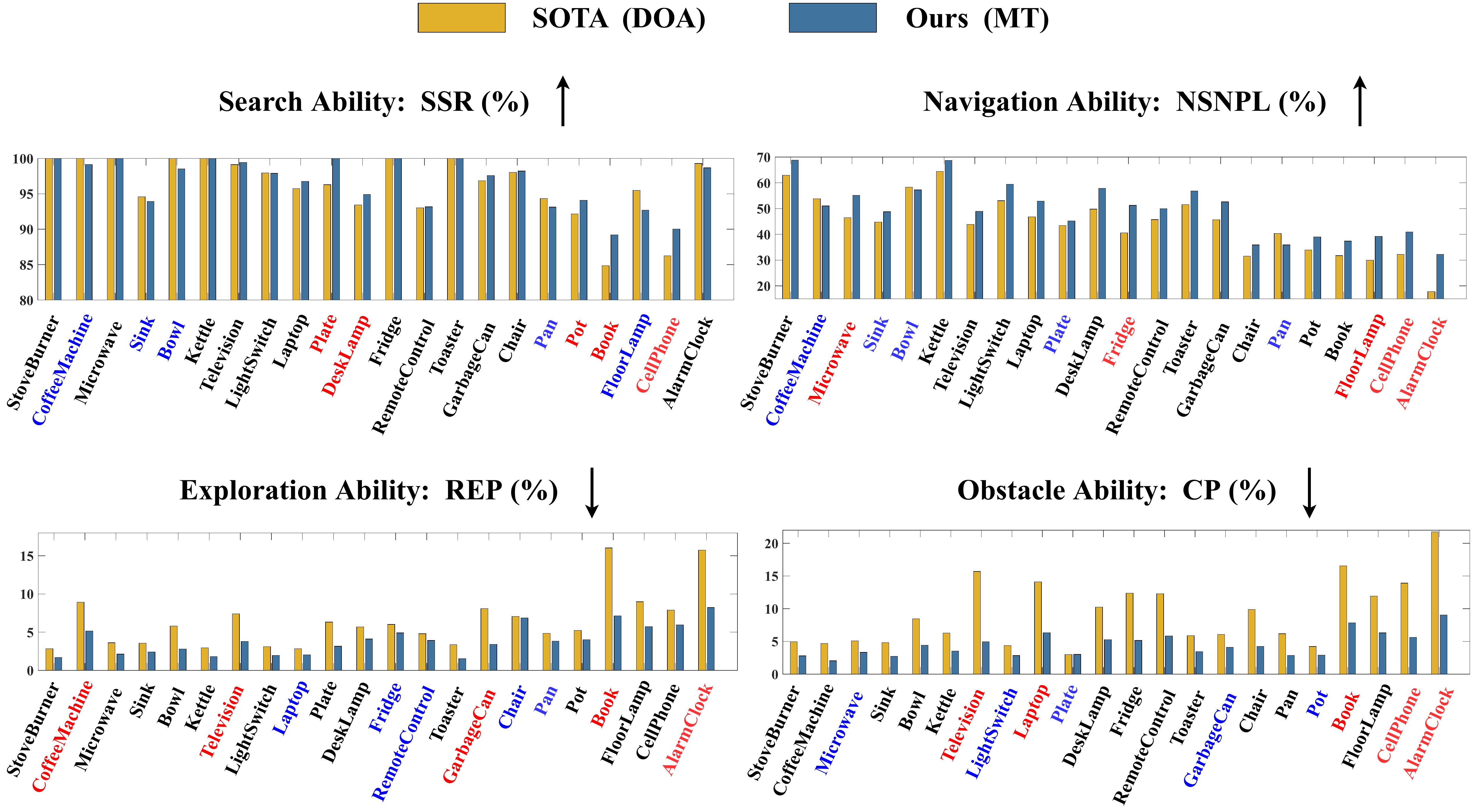}}
\caption{Comparison of the meta-ability metrics of the DOA method and the MT method at the level of targets. The
red and blue markers represent the targets with the best and worst
performance improvement of the MT method, respectively.}
\label{each_object_snrc}
\end{center}
\vskip -0.2in
\end{figure*}

\paragraph{Search Ability}
As observed in the SSR sub-figure of Figure~\ref{each_object_snrc}, the success rate of the agent in finding targets (ignoring distance) is already high, with the majority of targets having a search success rate of above 90\%. Among them, the search success rate of targets such as stove burner, kettle and fridge can even reach 100\%. This indicates that after the introduction of the associative mechanism, the algorithm's ability to search for target objects has indeed become very powerful. Our MT method borrows and simplifies some of the object attention allocation techniques from DOA in the encoding process of search thinking. Surprisingly, the search ability of MT not only does not decrease, but also improves in finding some targets (such as book and cellphone). There are two reasons for the phenomenon: (\romannumeral1) Although we decouple meta-abilities at the model level, various meta-abilities are mutually beneficial when completing tasks. In the search stage, the agent frequently fails because it is trapped in a local deadlock state and cannot escape. Therefore, if the agent has better environmental exploration ability and obstacle avoidance ability, the target searching can be easier. (\romannumeral2) Previously, models commonly wanted to implicitly abstract various meta-abilities through a kind of thinking. This would divert the attention of thinking, thereby making the inductive bias of meta-abilities less effective. After meta-abilities are decoupled, thinking is exclusive to a certain meta-ability and has a clearer learning direction, thus making the inductive bias more effective. 

\paragraph{Navigation Ability}
As observed in the NSNPL sub-figure of Figure~\ref{each_object_snrc}, the NSNPL metric exhibits a significant difference when searching for different targets compared to the SSR metric. In particular, for small objects such as alarm clock and cellphone located in complex environments, the navigation ability of the DOA method is inadequate. The advantages of the MT model are mostly reflected in these challenging targets. However, despite this, the MT method shows a slight decline in the NSNPL metric when the coffee machine, bowl, and pan are the target objects. These three objects are frequently found in kitchen environments, which are typically characterized by simple layouts and few obstacles. It can be inferred that the advantage of MT in simple scenarios is not as prominent as in complex scenarios.

\paragraph{Exploration Ability}
As observed in the REP sub-figure of Figure~\ref{each_object_snrc}, due to the lack of historical memory, the agent with DOA model frequently explores the same area repeatedly when looking for objects such as book and alarm clock. The repetitive exploration not only leads to wasted time, but also potentially disrupts the temporal reasoning logic of the model. The decoupled exploration ability shows significant improvement in addressing this issue.

\paragraph{Obstacle Ability}
As observed in the CP sub-figure of Figure~\ref{each_object_snrc}, the collision obstacle problem of the DOA method is very serious, and even repeatedly collides with the same obstacle. Objects such as laptop, book, and cellphone are commonly located in complex scene layouts, resulting in many highly constrained and difficult-to-navigate spaces for agents. The MT method’s obstacle ability is very obvious for the optimization of obstacle avoidance in such scenes. On the contrary, there are few dense obstacles next to objects such as plate, pot and light switch, so the demand for obstacle ability is less.

\begin{table*}[t]
\centering
\scriptsize
\setlength\tabcolsep{7.1pt}
\caption{The outcome of applying the DOA method with each object as the target on the AI2-Thor.}
\label{tab:each_object_DOA}
\begin{tabular}{c|cccccc|cccccc|c}
\hline
\multirow{2}{*}{MT} & \multicolumn{6}{c|}{ALL (\%)} & \multicolumn{6}{c|}{$L\ge 5$ (\%)} & \multirow{2}{*}{\begin{tabular}[c]{@{}c@{}}Episode\\ Length$\downarrow$\end{tabular}} \\
 & SR$\uparrow$ & SPL$\uparrow$ & SSR$\uparrow$ & NSNPL$\uparrow$ & REP$\downarrow$ & CP$\downarrow$ & SR$\uparrow$ & SPL$\uparrow$ & SSR$\uparrow$ & NSNPL$\uparrow$ & REP$\downarrow$ & CP$\downarrow$ &  \\ \hline
Alarm Clock & 31.42 & 17.65 & 99.28 & 17.77 & 15.74 & 21.74 & 25.58 & 15.67 & 99.22 & 14.87 & 15.84 & 23.85 & 37.35 \\
Book & 54.54 & 27.31 & 84.84 & 31.74 & 16.03 & 16.54 & 44.16 & 25.75 & 80.00 & 31.15 & 16.92 & 17.91 & 25.43 \\
Bowl & 88.33 & 51.55 & 100.00 & 58.33 & 5.82 & 8.46 & 84.84 & 53.59 & 100.00 & 57.48 & 6.72 & 9.26 & 12.85 \\
Cell Phone & 48.90 & 25.05 & 86.26 & 32.24 & 7.91 & 13.91 & 35.24 & 20.28 & 79.50 & 23.95 & 8.94 & 15.27 & 33.28 \\
Chair & 63.24 & 33.08 & 98.02 & 31.59 & 7.07 & 9.88 & 53.00 & 32.28 & 97.26 & 31.08 & 7.81 & 11.66 & 26.19 \\
Coffee Machine & 91.66 & 53.39 & 100.00 & 53.82 & 8.94 & 4.70 & 87.23 & 57.87 & 100.00 & 54.80 & 10.54 & 5.44 & 12.38 \\
Desk Lamp & 78.94 & 42.74 & 93.42 & 49.79 & 5.70 & 10.24 & 69.81 & 42.06 & 90.56 & 46.24 & 7.33 & 11.37 & 17.42 \\
Floor Lamp & 54.13 & 27.89 & 95.48 & 29.98 & 8.99 & 11.93 & 49.56 & 28.90 & 94.78 & 30.35 & 11.45 & 12.87 & 27.31 \\
Fridge & 77.77 & 42.66 & 100.00 & 40.59 & 6.02 & 12.38 & 72.09 & 45.06 & 100.00 & 44.12 & 8.31 & 14.07 & 21.18 \\
Garbage Can & 69.83 & 44.00 & 96.83 & 45.65 & 8.10 & 6.07 & 65.67 & 44.58 & 96.34 & 45.30 & 10.75 & 8.94 & 21.69 \\
Kettle & 87.50 & 55.95 & 100.00 & 64.39 & 2.94 & 6.28 & 85.71 & 59.42 & 100.00 & 64.07 & 4.77 & 7.81 & 18.00 \\
Laptop & 82.49 & 44.33 & 95.72 & 46.84 & 2.83 & 14.11 & 74.28 & 42.12 & 93.71 & 44.59 & 4.87 & 16.88 & 18.07 \\
Light Switch & 82.71 & 47.95 & 97.95 & 53.07 & 3.11 & 4.37 & 75.14 & 50.06 & 98.22 & 50.24 & 5.27 & 5.27 & 17.22 \\
Microwave & 91.23 & 46.97 & 100.00 & 46.50 & 3.62 & 5.10 & 87.50 & 52.92 & 100.00 & 51.16 & 5.12 & 6.26 & 15.49 \\
Pan & 62.26 & 35.40 & 94.34 & 40.34 & 4.85 & 6.18 & 56.82 & 30.99 & 93.18 & 38.83 & 4.91 & 7.31 & 24.92 \\
Plate & 79.63 & 41.12 & 96.29 & 43.41 & 6.32 & 3.01 & 72.97 & 42.23 & 94.59 & 44.35 & 7.04 & 3.77 & 17.57 \\
Pot & 56.86 & 30.42 & 92.15 & 33.91 & 5.23 & 4.23 & 41.93 & 23.56 & 87.09 & 25.75 & 6.15 & 4.92 & 25.98 \\
Remote Control & 77.52 & 43.35 & 93.02 & 45.75 & 4.80 & 12.30 & 69.76 & 40.91 & 90.69 & 41.89 & 5.82 & 13.24 & 17.07 \\
Sink & 90.17 & 43.43 & 94.57 & 44.80 & 3.57 & 4.83 & 80.34 & 44.86 & 90.34 & 47.74 & 5.08 & 5.71 & 13.46 \\
Stove Burner & 93.49 & 58.89 & 100.00 & 62.92 & 2.83 & 4.97 & 89.19 & 60.55 & 100.00 & 60.34 & 4.56 & 6.24 & 13.11 \\
Television & 84.21 & 44.84 & 99.12 & 43.88 & 7.40 & 15.69 & 79.10 & 48.43 & 98.76 & 49.81 & 8.22 & 15.31 & 18.07 \\
Toaster & 76.19 & 46.00 & 100.00 & 51.56 & 3.36 & 5.88 & 72.72 & 46.83 & 100.00 & 50.43 & 4.63 & 7.36 & 17.16 \\ \hline
\end{tabular}
\end{table*}

\begin{table*}[t]
\centering
\scriptsize
\setlength\tabcolsep{7.1pt}
\caption{The outcome of applying the MT method with each object as the target on the AI2-Thor.}
\label{tab:each_object_MT}
\begin{tabular}{c|cccccc|cccccc|c}
\hline
\multirow{2}{*}{MT} & \multicolumn{6}{c|}{ALL (\%)} & \multicolumn{6}{c|}{$L\ge 5$ (\%)} & \multirow{2}{*}{\begin{tabular}[c]{@{}c@{}}Episode\\ Length$\downarrow$\end{tabular}} \\
 & SR$\uparrow$ & SPL$\uparrow$ & SSR$\uparrow$ & NSNPL$\uparrow$ & REP$\downarrow$ & CP$\downarrow$ & SR$\uparrow$ & SPL$\uparrow$ & SSR$\uparrow$ & NSNPL$\uparrow$ & REP$\downarrow$ & CP$\downarrow$ &  \\ \hline
Alarm Clock & 46.02 & 31.14 & 98.66 & 32.22 & 8.22 & 9.03 & 41.21 & 29.10 & 96.83 & 30.17 & 11.92 & 12.15 & 22.41 \\
Book & 59.71 & 34.07 & 89.21 & 37.38 & 7.13 & 7.84 & 45.92 & 32.65 & 84.25 & 34.94 & 11.73 & 11.61 & 28.52 \\
Bowl & 86.40 & 57.26 & 98.52 & 57.24 & 2.80 & 4.42 & 82.11 & 59.16 & 100.00 & 51.61 & 3.85 & 4.37 & 10.05 \\
Cell Phone & 69.93 & 34.21 & 90.03 & 40.93 & 5.97 & 5.61 & 59.66 & 29.88 & 87.37 & 35.80 & 6.93 & 6.94 & 27.33 \\
Chair & 67.25 & 36.55 & 98.21 & 35.91 & 6.88 & 4.25 & 58.79 & 39.02 & 97.92 & 36.91 & 6.71 & 6.66 & 19.35 \\
Coffee Machine & 92.36 & 49.72 & 99.14 & 51.02 & 5.16 & 2.04 & 91.63 & 57.54 & 95.03 & 59.88 & 5.26 & 2.65 & 16.84 \\
Desk Lamp & 94.13 & 54.93 & 94.92 & 57.85 & 4.11 & 5.28 & 91.42 & 58.51 & 95.17 & 61.85 & 6.10 & 6.83 & 11.97 \\
Floor Lamp & 57.43 & 35.05 & 92.70 & 39.21 & 5.72 & 6.31 & 54.91 & 36.37 & 94.92 & 41.32 & 6.93 & 8.04 & 26.42 \\
Fridge & 88.55 & 48.32 & 100.00 & 51.28 & 4.91 & 5.16 & 91.05 & 53.54 & 100.00 & 56.01 & 6.15 & 6.82 & 16.26 \\
Garbage Can & 77.91 & 51.21 & 97.58 & 52.61 & 3.42 & 4.10 & 74.83 & 51.73 & 96.68 & 52.17 & 4.27 & 3.97 & 19.23 \\
Kettle & 93.72 & 61.83 & 100.00 & 68.72 & 1.81 & 3.54 & 92.99 & 62.04 & 100.00 & 69.43 & 2.05 & 3.85 & 18.84 \\
Laptop & 85.47 & 48.10 & 96.76 & 52.92 & 2.03 & 6.33 & 79.20 & 49.67 & 93.11 & 51.64 & 3.22 & 7.46 & 13.62 \\
Light Switch & 85.21 & 52.69 & 97.89 & 59.43 & 1.97 & 2.84 & 82.78 & 55.32 & 97.02 & 57.82 & 3.07 & 3.64 & 16.72 \\
Microwave & 96.49 & 55.08 & 100.00 & 55.13 & 2.14 & 3.34 & 96.25 & 61.47 & 99.78 & 59.55 & 3.14 & 3.16 & 11.62 \\
Pan & 49.30 & 29.94 & 93.12 & 35.90 & 3.83 & 2.87 & 46.74 & 28.81 & 93.62 & 31.02 & 4.17 & 3.15 & 23.24 \\
Plate & 80.05 & 45.33 & 100.00 & 45.23 & 3.16 & 3.02 & 73.85 & 42.13 & 100.00 & 43.00 & 3.24 & 3.00 & 18.01 \\
Pot & 53.81 & 31.75 & 94.10 & 39.04 & 4.02 & 2.91 & 36.19 & 83.42 & 84.10 & 26.31 & 4.38 & 2.54 & 28.74 \\
Remote Control & 79.27 & 46.22 & 93.17 & 49.95 & 3.92 & 5.83 & 72.71 & 47.17 & 90.25 & 48.79 & 5.06 & 6.75 & 17.13 \\
Sink & 89.45 & 48.79 & 93.92 & 48.77 & 2.41 & 2.72 & 78.34 & 55.69 & 91.00 & 55.67 & 4.85 & 4.07 & 10.62 \\
Stove Burner & 95.80 & 63.21 & 100.00 & 68.84 & 1.70 & 2.80 & 93.51 & 69.31 & 100.00 & 63.24 & 2.89 & 3.14 & 10.94 \\
Television & 81.41 & 45.98 & 99.43 & 48.96 & 3.80 & 4.96 & 75.06 & 48.33 & 96.20 & 51.99 & 5.74 & 5.95 & 13.62 \\
Toaster & 89.51 & 51.82 & 100.00 & 56.83 & 1.55 & 3.44 & 89.19 & 54.48 & 100.00 & 59.10 & 2.37 & 3.62 & 15.52 \\ \hline
\end{tabular}
\end{table*}

\section{Scene Level Experiment}
\label{sec:Appendix:E}
We discover that the contrasts between various scenes are quite clear from the analysis of each target object.  Therefore, we conduct experiments on the DOA method and our MT method in different scenes. We randomly select 1000 tasks (floorplan, target and initial state) from each scene and let the two methods complete these 1000 tasks simultaneously.  The results are illustrated in Table~\ref{tab:each_scene}.

\subsection{Result Analysis}
The agent depends on different navigation meta-abilities according to the variations in the scenes. In simple environments such as the kitchen and bathroom, the agent's reliance on intuition and search abilities alone is sufficient for efficient target navigation. The living room has a larger room area and a longer navigation path once the target is located. Therefore, the agent requires a stronger navigation ability to avoid repeated search in case of losing the target, as demonstrated by the improvement of 6.38/7.79 (ALL/$L\ge 5$, \%) in NSNPL with our MT model. Furthermore, the bedroom presents a greater challenge as not only is the search area larger, but the obstacles are also highly complex. With regard to these conditions, the agent requires stronger exploration and obstacle avoidance abilities, as evidenced by the significantly higher improvement in the REP and CP metrics compared to other scenes. 

Under the influence of the CV and NLP fields, many methods in embodied AI tasks tend to rely heavily on intuition ability to address all issues. To contradict this notion, it is important to recognize that complex, long-term decision-making tasks (e.g. visual language navigation (VLN), embodied question answering (EQA)) require more logical reasoning than simple intuition-based tasks (e.g. image classification, object detection).  If we solely focus on enhancing intuition ability, the algorithm will be limited to simple environments. Only agents endowed with multiple cognitive reasoning abilities could solve problems in complex environments. Our MAD paradigm provides a theoretical foundation and implementation guidance for the logical reasoning of agents.

\section{Meta-Ability Interpretability Analysis}
\label{Appendix:F}
In Sec.~\ref{sec:Meta-ability Qualitative Analysis} of the main text, we introduce the observation of meta-abilities behavior characteristics in navigation by visualizing the thinking activation at each step. However, some phenomena are interesting and unexpected. We will explain the causes of these phenomena in detail below.

\subsection{Switch of Rights Between Search Thinking and Navigation Thinking}
The dominance of search thinking and navigation thinking in decision making throughout most of the time makes the transition of power between them highly influential. As observed in Figure~\ref{visual_ai2thor} (a,b), upon sighting the target object, search thinking continues to dominate as long as the target object remains within the field of view. However, once the target object is out of sight, navigation thinking takes over as the primary decision-making leader. This phenomenon is a result of the model's long-term learning experience. When the target is in the field of view, the search thinking is able to locate it directly based on the target bounding box, which is more accurate than using the navigation thinking's memory knowledge to locate the target. Upon the target being lost from the field of view, search thinking becomes ineffective for target localization, making the ability of navigation thinking crucial. 

\subsection{Preference for Forward Actions in Exploration Thinking}
As seen in Figure~\ref{visual_ai2thor} (c), the activation intensity of rotation is less than that of the forward movement on exploration thinking. We explain this phenomenon with the following two reasons:

\begin{enumerate}[1)]
\setlength{\itemsep}{-0.1cm}
\item 
With the baseline model, the agent commonly becomes absorbed in rotation and rarely moves forward, which leads to a low success rate for objects in the distance. Therefore, we add a reward that promotes the agent's forward movement in order to alleviate this issue. Neural networks tend to choose the simplest way to fit the reward signal. Exploration thinking explicitly records the agent's historical path, thus learning to increase the probability of forward action output is the easiest. Consequently, the preference towards forward actions in exploration thinking is due to its task of enhancing the agent's forward action. 
\item 
From another perspective, we generally consider rotation as the most efficient method for acquiring environmental information, however, moving forward to acquire more detailed and accurate information is also crucial. 

\end{enumerate}

\begin{table*}[t]
\centering
\scriptsize
\setlength\tabcolsep{6pt}
\caption{Our MT method is compared with the DOA method in each scene. A scene is tested with 1000 episodes.}
\label{tab:each_scene}
\begin{tabular}{c|c|cccccc|cccccc|c}
\hline
\multirow{2}{*}{Scene} & \multirow{2}{*}{Method} & \multicolumn{6}{c|}{ALL (\%)} & \multicolumn{6}{c|}{$L\ge 5$ (\%)} & \multirow{2}{*}{\begin{tabular}[c]{@{}c@{}}Episode\\ Length$\downarrow$\end{tabular}} \\
 &  & SR$\uparrow$ & SPL$\uparrow$ & SSR$\uparrow$  & NSNPL$\uparrow$ & REP$\downarrow$ & CP $\downarrow$  & SR$\uparrow$ & SPL$\uparrow$ & SSR$\uparrow$  & NSNPL$\uparrow$ & REP$\downarrow$ & CP $\downarrow$  &  \\ \hline \hline
\multirow{3}{*}{Living Room} & DOA & 69.15 & 38.14 & 95.88 & 39.35 & 5.14 & 7.83 & 61.81 & 37.45 & 94.76 & 38.31 & 6.49 & 8.31 & 23.024 \\
 & MT & 73.47 & 42.96 & 95.27  & 45.73 & 3.81 & 5.07 & 67.88 & 44.38 & 93.72  & 46.10 & 4.83 & 5.80 & 20.394 \\
 & MT - DOA & 4.32 & 4.82 & -0.61 & 6.38 & -1.33 & -2.76 & 6.07 & 6.93 & -1.04 & 7.79 & -1.66 & -2.51 & -2.630 \\ \hline
\multirow{3}{*}{Kitchen} & DOA & 83.89 & 49.30 & 98.78 & 52.41 & 4.01 & 4.93 & 78.76 & 50.96 & 98.34 & 51.72 & 4.62 & 4.77 & 16.817 \\
 & MT & 87.62 & 55.84 & 98.51 & 55.30 & 3.37 & 3.91 & 82.75 & 56.66 & 97.94 & 55.91 & 3.65 & 3.72 & 16.172 \\
 & MT - DOA & 3.73 & 6.54 & -0.27 & 2.89 & -0.64 & -1.02 & 3.99 & 5.70 & -0.40 & 4.19 & -0.97 & -1.05 & -0.645 \\ \hline
\multirow{3}{*}{\textbf{Bedroom}} & DOA & 62.10 & 34.51 & 93.60 & 38.17 & 11.68 & 17.91 & 51.17 & 31.92 & 91.31  & 33.73 & 15.04 & 19.27 & 24.418 \\
 & MT & 75.70 & 44.07 & 96.35 & 45.81 & 6.29 & 6.53 & 65.04 & 43.62 & 93.65 & 44.22 & 8.54 & 7.84 & 20.548 \\
 & \textbf{MT - DOA} & \textbf{13.60} & \textbf{9.56} & \textbf{2.75}  & \textbf{7.64} & \textbf{-5.39} & \textbf{-11.38} & \textbf{13.87} & \textbf{11.70} & \textbf{2.34} & \textbf{10.49} & \textbf{-6.50} & \textbf{-11.43} & \textbf{-3.870} \\ \hline
\multirow{3}{*}{Bathroom} & DOA & 87.10 & 45.64 & 95.00 & 48.94 & 3.41 & 3.22 & 78.72 & 48.54 & 92.84 & 51.80 & 5.13 & 5.61 & 13.848 \\
 & MT & 89.05 & 51.17 & 95.43 & 51.76 & 3.62 & 3.28 & 82.33 & 55.03 & 93.51 & 56.07 & 5.66 & 5.56 & 13.250 \\
 & MT - DOA & 1.95 & 5.53 & 0.43 & 2.82 & 0.21 & 0.06 & 3.61 & 6.49 & 0.67 & 4.27 & 0.53 & 0.05 & -0.598 \\ \hline
\end{tabular}
\end{table*}

\end{document}